\documentclass[preprint,11pt]{article}

\usepackage{acl}
\usepackage{times}
\usepackage{latexsym}
\usepackage[T1]{fontenc}
\usepackage[utf8]{inputenc}
\usepackage{microtype}
\usepackage{inconsolata}
\usepackage{graphicx}

\usepackage{acronym}
\usepackage{svg}
\usepackage{amsmath}
\usepackage{rotating}
\usepackage[normalem]{ulem}
\useunder{\uline}{\ul}{}
\usepackage{xspace}
\usepackage{multirow}
\usepackage{placeins}
\usepackage{enumitem}
\usepackage{amssymb}
  \newcommand{\casestudy}[5]{%
    \begin{figure*}[t]
      \centering
      \begin{minipage}{0.95\textwidth}
        \textbf{Case.} #2
        \vspace{6pt}
  
        \textbf{LLM Response.}
        \begin{quote}
          \small\ttfamily #3
        \end{quote}
        \vspace{6pt}
  
        \textbf{Discussion.} #4
      \end{minipage}
      \caption{#1}
      \label{#5}
    \end{figure*}
  }

\title{Post-hoc Alignment of LLM-judges to Human Judgment Distribution}

\author{
    Sebastian Steindl$^1$ \quad Nikos Voskarides$^1$ \quad Alberto Gasparin$^2$ \quad Diego Marcheggiani$^1$ \\
    \\
    $^1$Amazon, Barcelona, Spain \quad $^2$Amazon, Berlin, Germany \\
    \texttt{\{sebstei,nvvoskar,marchegg\}@amazon.es}, \texttt{albgas@amazon.de}
  }

\begin{document}

\acrodef{NLG}{Natural Language Generation}
\acrodef{NLI}{Natural Language Inference}
\acrodef{NLP}{Natural Language Processing}
\acrodef{AutoEval}{Automatic Evaluation}
\acrodef{LLM}{Large Language Model}
\acrodef{IAA}{Inter-Annotator Agreement}
\acrodef{PHA}{Post-Hoc Alignment}
\acrodef{HLV}{Human Label Variation}
\acrodef{HJD}{Human Judgment Distribution}
\acrodef{MLE}{Maximum Likelihood Estimation}
\acrodef{CoT}{Chain-of-Thought}
\acrodef{AED}{Annotation Error Detection}
\acrodef{AutoEval}{Automatic Evaluation}
\acrodef{MLP}{Multilayer Perceptron}
\acrodef{ICL}{In-Context Learning}
\acrodef{LLMaJ}{LLM-as-a-judge}

\newcommand{\methodName}{\texttt{NAPHA}\xspace}

\maketitle

\begin{abstract}
The LLM-as-a-judge (LLMaJ) framework offers a cost-effective and reproducible solution for automatic evaluation.
However, current evaluation practices typically compare LLMaJ judgments against aggregated ground-truth labels, overlooking the valuable information contained in Human Label Variation (HLV).
Inspired by an increasing line of work that proposes to leverage HLV, we systematically study LLMaJ performance on predicting both a single, aggregated ground truth \emph{hard-label} and unaggregated \emph{soft-labels} that represent Human Judgment Distributions (HJD).
Our results across five diverse datasets reveal that while LLMs achieve near human-level performance at \emph{hard-label} prediction on most tasks, they exhibit poor performance when predicting \emph{soft-labels}.
To address this limitation, we propose \methodName (\textit{e\textbf{N}tropy-\textbf{A}ware \textbf{P}ost-\textbf{H}oc \textbf{A}lignment}), a simple yet effective lightweight post-hoc alignment method that matches the LLM distribution to the HJD by  first assigning an instance to a discrete entropy class and then routing it to specialized, trained alignment models.
We find that \methodName consistently improves \emph{soft-labels} prediction across base LLM models and datasets, with particularly strong gains on high-entropy instances where capturing diverse human perspectives is most critical.
We also show via oracle experiments  that improving entropy class prediction can substantially enhance \methodName's practical effectiveness.
\end{abstract}

\section{Introduction}
% Human annotation is a costly and time-consuming process, especially when it involves expert annotators.
%
Human labels are generally accepted as the gold-standard in NLP evaluation~\cite{clark-etal-2021-thats}.
During dataset construction, researchers oftentimes collect multiple annotations per instance using experts or crowd-workers~\cite{raykar2010learning, fabbri-etal-2021-summeval}.
Annotations are typically aggregated into a single ground truth label with some function, e.g., the mean, mode, or maximum.
This process of aggregating annotations eliminates disagreement in favor of having a single ground truth. 
In fact, low disagreement among annotators is commonly used to suggest high annotation quality~\cite{clark-etal-2023-seahorse}, sometimes leading to a harmonization step, where annotators re-annotate instances they disagreed on~\cite{fabbri-etal-2021-summeval}.

\begin{figure}[]
    \centering
    \includegraphics[width=\linewidth]{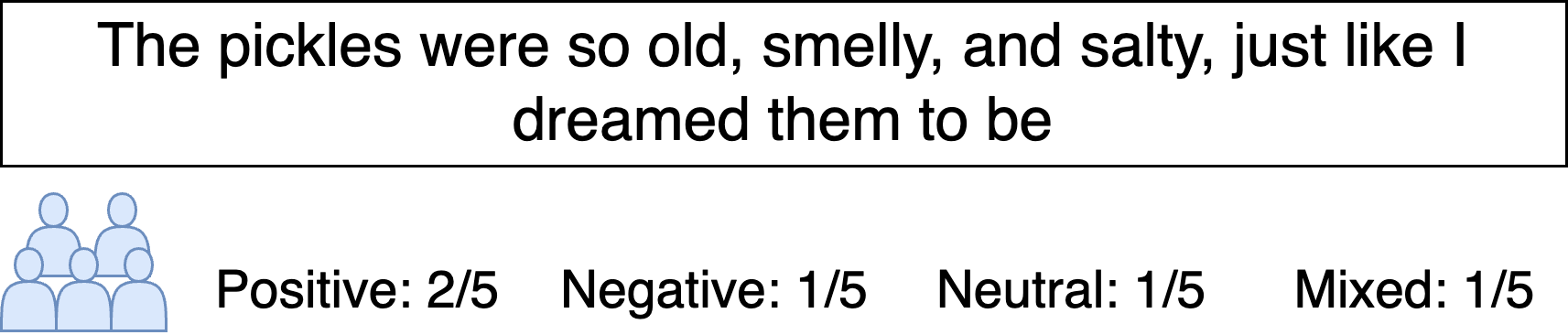}
    \caption{Example (id: r2-0011275) from the DynaSent dataset~\cite{potts-etal-2021-dynasent}. The sentence begins with adjectives that are mostly negatively connoted, but ends on a positive note. This shift in expressed sentiment leads to relatively high disagreement among the five annotators; it is unclear whether the ending is meant to be ironic or not. }
    \label{fig:example_disagreement}
\end{figure}

Recent work has questioned the assumption that a single ground truth label invariably exists~\cite{de-marneffe-etal-2012-happen,plank-etal-2014-linguistically,aroyoWelty2015,pavlick-kwiatkowski-2019-inherent}.
This has culminated in a line of work that proposes not to eliminate, but instead leverage \ac{HLV} in annotations \cite{plank-2022-problem,cabitza2023toward}. 
Figure \ref{fig:example_disagreement} shows an example of how plausible \ac{HLV} can manifest in human annotations.
The \ac{HLV} paradigm is based on the idea that different labels can be simultaneously correct, as they represent different annotator backgrounds or task ambiguity \cite{plank-2022-problem,cabitza2023toward}.
Thus, in many tasks, such as toxicity detection or sentiment classification, there can be irreconcilable variation as a form of aleatoric uncertainty in the annotations  \cite{plank-2022-problem,cabitza2023toward}. This occurs even for supposedly objective tasks, such as those in the medical domain \cite{schaekermann2019understanding,cabitza2019elephant}.
Leveraging HLV in annotations can therefore: i) better reflect reality of irreconcilable variation, uncertainty and decision making \cite{cabitza2023toward}, ii) improve fairness by considering minority positions \cite{noble2012minority,baan-etal-2024-interpreting}, and even iii) achieve better performance and generalization \cite{washington2021training,uma2020case,peterson2019human,uma2021learning}. 
Recently, \ac{LLMaJ} has emerged as a promising framework for cost-effective and reproducible automatic evaluation~\cite{LLM_judge_NLG_survey}.
As we move towards leveraging \ac{HLV}, automatic evaluation in this context becomes increasingly important.
Therefore, in this paper, we investigate the capabilities of state-of-the-art \ac{LLMaJ}  models to produce the aggregated ground truth \emph{hard-label}, and the \emph{soft-labels} ground truth, i.e., the \ac{HJD} in the context of \ac{HLV}.
We conduct our study on five datasets that represent typical \ac{NLG} and classification tasks.
We find that, while \ac{LLMaJ} achieve near human-level performance on \emph{hard-label} prediction on some tasks, it performs poorly on \emph{soft-labels} prediction.
To address this limitation, we propose \methodName (\textit{e\textbf{N}tropy-\textbf{A}ware \textbf{P}ost-\textbf{H}oc \textbf{A}lignment}), a simple yet effective lightweight post-hoc alignment method that, given the \emph{soft-labels} predictions of a given instance, first assigns the instance to a discrete entropy class and then routes it to specialized, trained alignment models.
%
% We find that \methodName improves performance on \emph{soft-labels} prediction.
We demonstrate that \methodName consistently improves \emph{soft-labels} prediction across models and datasets.
\methodName can be applied in scenarios where the user expects that different human perspectives could lead to multiple labels being true at the same time, most notably subjective tasks.

Our work is complementary to \citet{elangovan2025beyond}, who demonstrate that in scenarios of significant HLV, \ac{LLMaJ} can reach the same correlation to the human annotations as the human-to-human comparison.
They argue that this is misleading, as it is a consequence of the human disagreement.
In our work, we instead follow the paradigm of \ac{HLV}: we consider disagreement as a valuable signal and thus propose \methodName as a technical solution to improve the capability of \ac{LLMaJ} to predict the \emph{soft-labels}.

In summary, our contributions are:
\begin{itemize}
    \item We investigate the performance of \ac{LLMaJ} using state-of-the-art LLMs for predicting \emph{hard-label} and \emph{soft-labels} on a broad range of tasks.
    \item We showcase that, by default, \ac{LLMaJ} can accurately predict the \emph{hard-label} at human-level performance, but not the \emph{soft-labels}. We validate this finding on multiple datasets with several state-of-the-art LLMs.
    \item We propose \methodName, a post-hoc alignment method that improves  \emph{soft-labels} prediction compared to the base \ac{LLMaJ} models across models and datasets. Through extensive experimentation, oracle studies and ablations we demonstrate \methodName's performance and identify potential for further improvements.
\end{itemize}

\section{Background and Related Work}
\subsection{Automatic Evaluation and \ac{LLMaJ}}
\ac{AutoEval} focuses on creating evaluation algorithms that can alleviate annotation cost and improve reproducibility \cite{LLM_judge_NLG_survey}. Early attempts include metrics such as ROUGE~\cite{lin-2004-rouge} and later BERTScore~\cite{BERTScore}, which have low correlation with human annotations \cite{gehrmann2023repairing}.
Recently, the \ac{LLMaJ} framework \cite{LLMJudge,wang-etal-2023-chatgpt} has gained considerable attention. By using a prompt-engineered \ac{LLM} as the evaluation function, one can obtain reproducible, cheap judgments, that have shown better alignment to human annotations than prior automatic metrics \cite{liu-etal-2024-hd,lee-etal-2025-checkeval}.
%
% \citet{LLM_judge_NLG_survey} discusses current status and challenges of \ac{LLMaJ}.

\subsection{Human Label Variation (HLV)}
The terms \ac{HLV} \cite {plank-2022-problem} and \textit{perspectivism} \cite{cabitza2023toward} describe a paradigm in which the existence of irreconcilable disagreement in annotations is accepted and even attempted to leverage~\cite{weber-genzel-etal-2024-varierr,hong-etal-2025-litex,ramponi-etal-2025-fine,gruber-etal-2025-revisiting,chen-etal-2025-threading}.\footnote{We use the term HLV throughout this paper.}
As such, it is also related to the call for pluralistic model alignment by \citet{sorensenRoadmapPluralistic}.
In their taxonomy, our study falls into the category of distributionally pluralistic models. 
A common way to consider HLV is by using soft-labels, i.e., a distribution across the possible labels, instead of a single-ground truth, which we adopt in our work.

%\citet{hong-etal-2025-litex} build upon the idea of HLV and investigate within-label variation, i.e., annotators giving the same label, but for different reasons.
%\citet{weber-genzel-etal-2024-varierr} provide annotators' explanations on an \ac{NLI} task, and \citet{ramponi-etal-2025-fine} publish a dataset on fallacy detection, that considers HLV by design.
%
In general, not all HLV is valid, since annotation errors can still exist in this context and thus lead to label noise. Discerning between noise and valid HLV is not straightforward, but an active research question \cite{klie2023AED,weber-plank-2023-activeaed,ivey-etal-2025-nutmeg,nahum-etal-2025-llms}.
Our focus is not on disentangling error from valid variation, and we thus consider all HLV in the datasets to be valid.
By attempting to consider all opinions deriving from human annotations, leveraging HLV becomes important in relation to efforts towards accounting for diversity in terms of background and culture
\cite{santurkar2023whose,kotek2023gender,mukherjee-etal-2023-global,durmus2024towards,adilazuarda-etal-2024-towards,liu-etal-2025-culturally}.

\subsection{Calibration Methods} The problem of calibration, which describes the process of aligning the model prediction with its accuracy, has previously been treated by methods that assume access to the model architecture \cite{cho2024tilt,tao2025feature}. In the field of computer vision, parameterized temperature scaling \cite{tomani2022parameterized} has been proposed as an evolution of standard temperature scaling. For language models, it has been discussed, e.g., by \citet{jiang-etal-2021-know} and \citet{zheng2023large}. Strictly speaking, the post-hoc alignment of the LLM-judge to the HJD is not a calibration problem, since this distribution does not reflect uncertainty of choosing one class, but that different classes can be true at the same time. Therefore, the model uncertainty would instead be the uncertainty about the distribution. Thus, while our work relates to calibration literature, we instead focus on the alignment to the HJD, regardless of the model's uncertainty.

\section{Problem Statement}
\label{sec:problem-statement}
Let $T$ be an evaluation task with a set of $n$ possible discrete labels $\mathcal{A} = \{a_1, \dots, a_n\}$. 
An example evaluation task $T$ is to classify the coherence of a single generated text into one of $n$ grades.
A dataset instance is created by collecting $m$ human labels for $T$: $\mathbf{r} = (r_1, \dots, r_m)$ with all $r_i \in \mathcal{A}$.

We study two evaluation settings. %
In the first setting, we obtain a ground truth \emph{hard-label} $y$ by combining the human labels $\mathbf{r}$ using an aggregation function $agg(\cdot)$, which can be majority-voting or the mean.
In this setting, a predicted \emph{hard-label} $\hat{y}$ is evaluated against the ground truth \emph{hard-label} $y$ using the F1-score or a correlation metric across instances.
The second evaluation setting considers \ac{HLV}, which is the focus of this work. 
Here, we obtain ground truth \emph{soft-labels} by using a categorical distribution $\mathbf{y} = (y_1, \dots, y_n)$ that represents the probability over the possible discrete labels $\mathcal{A}$. We estimate $\textbf{y}$ via maximum likelihood using the observed human labels $\mathbf{r}$.
In this setting, the predicted \emph{soft-labels} $\mathbf{\hat{y}}$ are evaluated against the ground truth \emph{soft-labels} using a distance metric $dist(\mathbf{\hat{y}}, \mathbf{y})$.\footnote{We discuss evaluation metrics in more detail in Section~\ref{sec:metrics}.}

\section{Predicting \emph{soft-labels}}
% \todo{our framework builds on top base models.}
In this section we first describe base \ac{LLMaJ} models for predicting \emph{soft-labels} that do not use explicit alignment to HJD (Section~\ref{sec:base-models}), and then describe \methodName, our proposed post-hoc alignment method that builds on top of the base models (Section~\ref{sec:napha}).
Note that, for the more commonly used setting of predicting a \emph{hard-label}, we use standard \ac{LLMaJ} with a prompt constructed for each specific task.\footnote{Prompts are provided in Appendix~\ref{sec:appdx_prompts}.}

\subsection{Base \ac{LLMaJ} Models}
\label{sec:base-models}
We study three different ways for predicting the \emph{soft-labels} $\mathbf{\hat{y}}$ with an \ac{LLM}, to which we refer as the base models.
%
%Each base model uses a ``backbone'' LLM.

Our first base model builds on the SimulatedAnnotators (SimAnn) \cite{simulatedAnnots} approach; it samples a response multiple times from the LLM to predict the soft labels $\mathbf{\hat{y}}$.
Also, to increase output diversity, we set the temperature to $t=1$ and provide different \ac{ICL} examples in every run~\cite{brown2020language,sanh2022multitask,dong-etal-2024-survey}.
%We extract the soft-label distribution as the \ac{MLE} of the individual simulated annotators' predictions. 
We use ten simulated annotators with five \ac{ICL} examples each.

Next, we investigate base models that directly prompt the LLM to predict the \emph{soft-labels}~\cite{mielke-etal-2022-reducing,xiong2024can,geng-etal-2024-survey}.
We construct two base models: one with \emph{hard-label} \ac{ICL} examples in the prompt, which we name SLP-HE (Soft Label Prediction with Hard label examples), and one with \emph{soft-labels} \ac{ICL} examples in the prompt, which we name SLP-SE (Soft Label Prediction with Soft label Examples).
We provide more details in Appendix \ref{sec:appdx_prompts}.

\subsection{Post-hoc alignment with \methodName}
\label{sec:napha}

\begin{figure*}[t]
    \centering
    \includegraphics[width=0.75\linewidth]{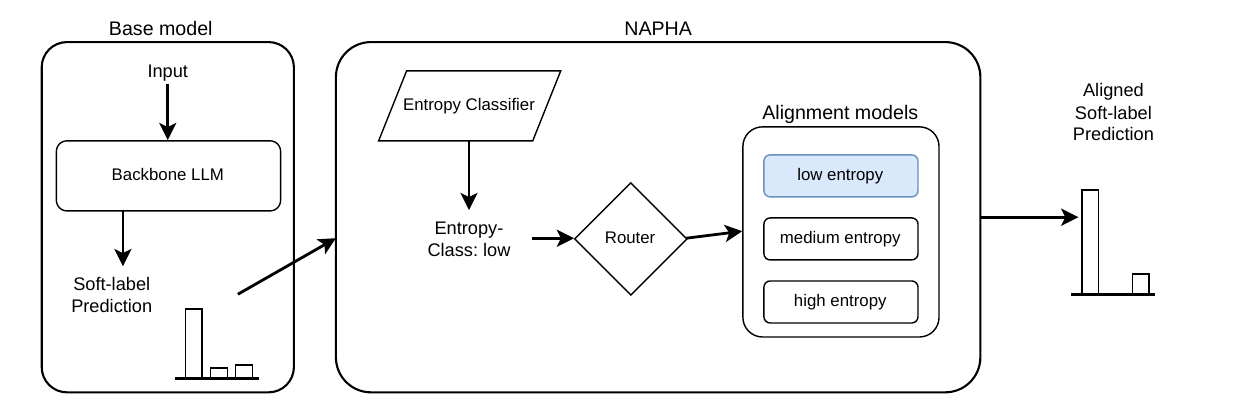}
    \caption{Visualization of \methodName. The base model \ac{LLMaJ} predicts \emph{soft-labels} for a given instance. Then, the Entropy Classifier outputs an entropy class that is used to route the predicted \emph{soft-labels} to the corresponding alignment model specialized in that entropy class, which in turn outputs the final (aligned) predicted \emph{soft-labels}.}
    \label{fig:methodFig}
\end{figure*}

As we will show in Section \ref{sec:results}, independently of which base model is used, LLMs perform poorly at predicting \emph{soft-labels}.
To mitigate this problem, we propose \textit{e\textbf{N}tropy-\textbf{A}ware \textbf{P}ost-\textbf{H}oc \textbf{A}lignment} (\methodName), a simple yet effective post-hoc alignment method that operates on top of a base \ac{LLMaJ} model.
A theoretical motivation of our approach is described in Appendix~\ref{sec:formalization}.
Motivated by the observation that LLM-HJD mis-alignment patterns differ substantially across entropy strata, \methodName first assigns an instance to a discrete  entropy class given prediction $\hat{\mathbf{y}}$, which was obtained by the base model.
Subsequently, depending on the predicted entropy class $c$, \methodName routes $\hat{\mathbf{y}}$ to the corresponding trained alignment model $M_{c}$.
The output of $M$ is the final (aligned) \emph{soft-labels} prediction.
We illustrate the inference pipeline in Figure \ref{fig:methodFig}.

Next we describe \methodName in more detail.
Given the predicted \emph{soft-labels} $\mathbf{\hat{y}}$ obtained by the base model, we first compute the Shannon entropy over each instance in a dataset:
$
H(\mathbf{\hat{y}}) = -\sum_{j=1}^{n} \hat{y}_j \log \hat{y}_j
$.
Then, we assign each instance to an entropy class $c$ based on the entropy terciles:
\begin{equation*}
c = \begin{cases}
\text{low} & \text{if } H(\hat{\mathbf{y}}) \leq Q_{1} \\
\text{medium} & \text{if } Q_{1} < H(\hat{\mathbf{y}}) \leq Q_{2} \\
\text{high} & \text{if } H(\hat{\mathbf{y}}) > Q_{2}
\end{cases}
\end{equation*}
where $Q_{1}$ and $Q_{2}$ are the first and second terciles of the entropy distribution in a dataset, respectively, and $c \in \{\text{low}, \text{medium}, \text{high}\}$.\footnote{Preliminary experiments with using only two entropy classes (low and high) did not lead to performance improvements.}
In order to account for the fact that LLM mis-alignment patterns differ substantially across entropy strata, we train one specialized alignment model per entropy class $c$.
Thus, after assigning an entropy class to an instance, we route its predicted \emph{soft-labels} to one of the specialized alignment models according to their assigned entropy class $c$.
We define the alignment models as follows:
\begin{equation*}
M(c, \cdot) = \begin{cases}
M_{\text{low}}(\cdot) & \text{if } c = \text{low} \\
M_{\text{medium}}(\cdot) & \text{if } c = \text{medium} \\
M_{\text{high}}(\cdot) & \text{if } c = \text{high}
\end{cases}
\end{equation*}

The alignment models take the predicted \emph{soft-labels} $\mathbf{\hat{y}}$ as input and apply a non-linear transformation on them. They are trained to better match the ground-truth \emph{soft-labels}:
$
\mathbf{y} = M(c, \hat{\mathbf{y}})$.
As we will show in Section ~\ref{sec:ablations}, \methodName does not need a large amount of training data to improve performance.

The entropy class of each instance is decided by binning the \emph{soft-labels} predictions into three entropy classes based on their entropy $H$.\footnote{Preliminary experiments on prompting for entropy classes showed no benefit over calculating the entropy-level based on the predicted \emph{soft-labels}. We thus only focus on the latter case.}
In Section~\ref{sec:results-soft-labels} we discuss results when using the oracle entropy class, which is obtained by calculating the entropy on the ground truth \emph{soft-labels}, which we consider as an upper bound for \methodName.

%\textbf{Entropy Classifier}. 

In stratifying instances based on their entropy, our approach is similar in spirit with \citet{jolly-etal-2021-ease} and also implements the recommendation by \citet{elangovan2025beyond} to stratify LLM predictions by human uncertainty.
We highlight that \methodName does not require access to LLM internals but only to the output tokens.

% \subsection{Training \methodName alignments}
\paragraph{Training \methodName alignment models}
\methodName alignment models are lightweight and add negligible computational overhead. They are built on a simple \ac{MLP} architecture that consists of a single hidden layer with a ReLU activation. 
The dimension of both the input and output layers is $n$ (the number of possible discrete labels $\mathcal{A}$), and the dimension of the hidden layer is $n \times 2$.
We apply a softmax function to the output layer, which allows us to treat the outputs as probabilities.
We employ KL-divergence as the loss function and apply weight decay regularization.
Note that we compared to different calibration methods as alignment models, none of which outperformed the \ac{MLP} architecture:  a linear transformation, temperature scaling \cite{guo2017calibration}, Dirichlet calibration \cite{kull2019beyond} and Parameterized Temperature Scaling \cite{tomani2022parameterized} (see Table \ref{tab:ablation_otherCalibrationModels}).

\section{Experimental Setup}

\subsection{Datasets}
In order to study model performance in the context of HLV, we use five datasets on different subjective evaluation tasks, which either have score-based output (e.g., 1-5) or selection-based output (i.e., classification)~\cite{li-etal-2025-generation}.
All datasets have multiple human annotations per sample, and different sources of HLV (discussed in Appendix~\ref{sec:analysis-of-hjd}):

   \begin{description}[style=unboxed,leftmargin=0pt,nosep]
     \item[SummEval \cite{fabbri-etal-2021-summeval}] A news summarization dataset; each instance
   is annotated by three experts on multiple criteria.
     \item[TopicalChat \cite{gopalakrishnan2019topical}] A human-human conversation dataset with
   annotations for next utterances, each annotated by three crowd-workers on multiple criteria.
     \item[ChaosNLI \cite{nie-etal-2020-learn}] An \ac{NLI} dataset constructed by re-annotating
   high-disagreement SNLI \cite{bowman-etal-2015-large} samples. Each sample has 100 crowd-worker
   annotations.
     \item[DynaSent \cite{potts-etal-2021-dynasent}] A sentiment classification dataset
   (positive/negative/neutral). We use the more challenging ``round 2'' adversarial subset.
     \item[Anecdotes \cite{anecdotesData}] A Reddit dataset with ethical ``who is in the wrong''
   questions. Annotations are extracted from votes; we use instances with $\geq$15 votes.
   \end{description} 

For computational efficiency, we limit the DynaSent and Anecdotes to 1500 examples each. To study how annotator disagreement affects LLM-judge performance, we stratify sampling by disagreement level: we calculate
binary entropy across soft-labels and sample 500 examples from each quantile (low, medium, and high).
For all datasets, we report main results with a 20/80 train/test split to reflect realistic limited data availability for training the alignment model.\footnote{We observe consistent trends when using different splits, as shown in Table \ref{tab:ablation_trainingData} in Appendix \ref{sec:appx_ablations}.
}

\subsection{Evaluation Metrics}
\label{sec:metrics}

% \paragraph{Hard-labels} 
% First, we measure how well the backbone LLM is aligned to the aggregated ground-truth. 
%For the SummEval and TopicalChat dataset, we aggregate by taking the mean of the annotations, as is common practice in such ordinal rating tasks.
%For these datasets, we report the correlation as measured by Kendall's $\tau$ and the Advantage Probability (AP) from the \textit{alt-test} \cite{calderon-etal-2025-alternative}.
%
% Here we describe the evaluation metrics used to measure how well model predictions approximate the ground truth \emph{hard-label} or the \emph{soft-labels}.

\paragraph{Metrics for predicting the \emph{hard-label}}
For ChaosNLI, Anecdotes and DynaSent, the ground-truth \emph{hard-label} is defined as the majority label. 
We estimate the average human performance by sampling the human prediction according to the label distribution and comparing it to the majority label.
We measure the alignment to the \emph{hard-label} as the macro-average F1-score. 
For the ordinal rating tasks SummEval and TopicalChat, we follow common practice and define the \emph{hard-label} as the average rating. 
We report correlation as measured by Kendall's $\tau$, and the Advantage Probability (AP) from the alt-test \cite{calderon-etal-2025-alternative}.
We estimate human performance by modifying the bootstrapping approach from \citet{bavaresco-etal-2025-llms} to a leave-one-out bootstrapping; we use this to estimate the average correlation from one annotator to the leave-one-out \emph{hard-label}. More details are given in Appendix \ref{appdx:leaveOneOut}.
% We prompt the model to output a single label.

\paragraph{Metrics for predicting \emph{soft-labels}} In order to measure the distance between the ground truth and predicted soft-labels, we use the Distribution Calibration Error (DistCE), which expresses the maximum discrepancy in probability between the human and model-predicted soft-label distribution, over all possible events \cite{baan-etal-2022-stop}.
We also use the Jensen-Shannon Distance (JSD) \cite{jensenShannonMetric}.

\subsection{Implementation Details}
Prior work links model size to soft-label quality \cite{madaan-etal-2025-lost}, with mixed distillation results \cite{chen-etal-2024-seeing}.
Thus, our main results are focused on a state-of-the-art closed-weights LLM, Claude-4-Sonnet \cite{claude4}.\footnote{Note that, even though access to logits or internal model state is not always possible for closed-weights LLMs \cite{geng-etal-2024-survey}, \methodName can be used both with closed-weights and open-weights LLMs.}
We also provide results with GPT-OSS-120B~\cite{agarwal2025gpt} and Qwen3-32B \cite{yang2025qwen3} in Appendix \ref{sec:appdx_otherModels}, where we show that the choice of the LLM does not change the main conclusions.
We perform one LLM inference run with temperature $t = 0$. Main results are reported as the mean and standard error derived from the 95\% confidence interval over 20 independent post-hoc alignment runs to measure the stability of the post-hoc alignment with \methodName.

\section{Results and Discussion}
\label{sec:results}

\subsection{Predicting the \emph{hard-label}}

Tables \ref{tab:f1_scores} and \ref{tab:hard-label_sme_topical} show that, overall, the \ac{LLMaJ} performs close to human-level when predicting hard labels, across all datasets.\footnote{The average human performance is estimated by repetitively sampling a prediction based on the \emph{soft-labels}.}
% we find that the judge-LLM is able to perform on human-level . 
%especially on the selection-based ChaosNLI, Anecdotes, and DynaSent, and the score-based task TopicalChat.
On the ChaosNLI dataset the \ac{LLMaJ} outperforms the average human annotator, achieving an F1 score of 0.99, 0.81 and 0.61 for low, medium, and high entropy, compared to the human 0.93, 0.74, 0.53.
The Anecdotes dataset is more challenging: the LLM achieves 0.48, 0.48, 0.34 compared to the human 0.50, 0.42, 0.35. 

On SummEval, a gap remains between human and LLM performance: the average LLM Kendall's $\tau$ is 0.480 compared to the human score of 0.542. This likely reflects
the dataset creators' efforts to minimize annotator disagreement. Higher disagreement makes it easier for LLMs to achieve human-level performance, as humans are more
likely to deviate from the majority label~\cite{elangovan2025beyond}.

\begin{table}[t]
\centering
\resizebox{\linewidth}{!}{%
\begin{tabular}{lcccccc}
\hline
 & \multicolumn{2}{c}{Low $H$} & \multicolumn{2}{c}{Medium $H$} & \multicolumn{2}{c}{High $H$} \\
Dataset & LLM & Human & LLM & Human & LLM & Human \\ \hline
ChaosNLI & 0.99 & 0.93 & 0.81 & 0.74 & 0.61 & 0.53 \\
Anecdotes & 0.48 & 0.50 & 0.48 & 0.42 & 0.34 & 0.35 \\
DynaSent & 0.95 & 1.00 & 0.81 & 0.80 & 0.40 & 0.43 \\ \hline
\end{tabular}
} % end resizebox
\caption{Macro-Average F1 Scores for \emph{hard-label} predictions from Claude-4 and humans, stratified by entropy class.}
\label{tab:f1_scores}
\end{table}

\begin{table*}[th]
\centering
\resizebox{\linewidth}{!}{%
\begin{tabular}{llcccccccccc}
\hline
\multirow{2}{*}{\bf Dataset} & \multirow{2}{*}{\bf Predictions} & \multicolumn{2}{c}{Coherence} & \multicolumn{2}{c}{Fluency} & \multicolumn{2}{c}{Relevance} & \multicolumn{2}{c}{Consistency} & \multicolumn{2}{c}{Average} \\
 & \ & $\tau$ $\uparrow$ & AP $\uparrow$ & $\tau$ $\uparrow$ & AP & $\tau$ $\uparrow$ & AP & $\tau$ $\uparrow$ & AP $\uparrow$ & $\tau$ $\uparrow$ & AP $\uparrow$ \\ \hline
\multirow{2}{*}{SummEval} & LLM & 0.488 & 0.759 & 0.397 & 0.233 & 0.401 & 0.684 & 0.635 & 0.874 & 0.480 & 0.638 \\
 & Humans & 0.503 & - & 0.559 & - & 0.381 & - & 0.726 & - & 0.542 & - \\ \hline
 &  & \multicolumn{2}{c}{Naturalness} & \multicolumn{2}{c}{Coherence} & \multicolumn{2}{c}{Engagingness} & \multicolumn{2}{c}{Groundedness} & \multicolumn{2}{c}{Average} \\ \hline
\multirow{2}{*}{TopicalChat} & LLM & 0.554 & 0.794 & 0.634 & 0.887 & 0.598 & 0.824 & 0.756 & 0.972 & 0.636 & 0.870 \\
 & Humans & 0.435 & - & 0.528 & - & 0.544 & - & 0.731 & - & 0.559 & - \\ \hline
\end{tabular}
} % end resizebox
\caption{Results on \emph{hard-label} prediction from Claude-4 for the SummEval and TopicalChat dataset. Reporting Kendall's $\tau$ and AP \cite{calderon-etal-2025-alternative}. The $\tau$ values for the Human performance are obtained via Leave-one-out bootstrapping (cf. Appendix \ref{appdx:leaveOneOut}).}
\label{tab:hard-label_sme_topical}
\end{table*}

\subsection{Predicting \emph{soft-labels}}
\label{sec:results-soft-labels}

\begin{table*}[t]
\centering
\resizebox{\linewidth}{!}{%
\begin{tabular}{lc|ccc|ccc|ccc}
\hline
\multirow{3}{*}{\bf Dataset} & \bf Base model & \multicolumn{3}{c|}{SLP-HE} & \multicolumn{3}{c|}{SLP-SE} & \multicolumn{3}{c}{SimAnn} \\
 & \bf Alignment & - & NAPHA & NAPHA & - & NAPHA & NAPHA & - & NAPHA & NAPHA \\
 & \bf Entropy Label Routing & - & predicted & oracle & - & predicted & oracle & - & predicted & oracle \\ \hline
\multirow{2}{*}{Anecdotes} & DistCE $\downarrow$ & 0.299 & $0.271$ (0.001) & $0.175$ (0.002) & 0.299 & $0.272$ (0.002) & $0.172$ (0.001) & 0.325 & $0.293$ (0.001) & $0.172$ (0.001) \\
 & JSD $\downarrow$ & 0.326 & $0.292$ (0.001) & $0.227$ (0.002) & 0.326 & $0.294$ (0.001) & $0.224$ (0.001) & 0.385 & $0.308$ (0.001) & $0.225$ (0.001) \\ \hline
\multirow{2}{*}{ChaosNLI} & DistCE $\downarrow$ & 0.221 & $0.187$ (0.002) & $0.158$ (0.002) & 0.200 & $0.174$ (0.001) & $0.153$ (0.002) & 0.247 & $0.190$ (0.002) & $0.144$ (0.001) \\
 & JSD $\downarrow$ & 0.259 & $0.213$ (0.001) & $0.194$ (0.001) & 0.230 & $0.198$ (0.001) & $0.186$ (0.002) & 0.305 & $0.216$ (0.001) & $0.181$ (0.001) \\ \hline
\multirow{2}{*}{DynaSent} & DistCE $\downarrow$ & 0.268 & $0.263$ (0.003) & $0.158$ (0.002) & 0.272 & $0.265$ (0.002) & $0.172$ (0.004) & 0.296 & $0.294$ (0.002) & $0.158$ (0.004) \\
 & JSD $\downarrow$ & 0.313 & $0.306$ (0.002) & $0.236$ (0.002) & 0.330 & $0.305$ (0.001) & $0.249$ (0.003) & 0.367 & $0.323$ (0.001) & $0.238$ (0.003) \\ \hline
\multirow{2}{*}{SummEval} & DistCE $\downarrow$ & 0.433 & 0.303 & 0.262 & 0.371 & 0.303 & 0.263 & 0.562 & 0.305 & 0.258 \\
 & JSD $\downarrow$ & 0.465 & 0.356 & 0.334 & 0.453 & 0.358 & 0.335 & 0.560 & 0.360 & 0.332 \\ \hline
\multirow{2}{*}{TopicalChat} & DistCE $\downarrow$ & 0.363 & 0.349 & 0.301 & 0.374 & 0.363 & 0.312 & 0.374 & 0.349 & 0.282 \\
 & JSD $\downarrow$ & 0.404 & 0.376 & 0.356 & 0.424 & 0.385 & 0.363 & 0.433 & 0.375 & 0.342 \\ \hline
\end{tabular}
} % end resizebox
\caption{Results on soft-label prediction across all entropy classes. We report mean and standard error from 20 runs. For SummEval and TopicalChat we show the average across all evaluation criteria and thus do not show the standard error. Note that the oracle entropy class is generally unavailable but we use it as an upper bound in performance. Detailed results can be found in Appendix \ref{sec:appdx_fullResults}.}
\label{tab:result_soft_overview}
\end{table*}

\begin{figure*}[t]
\centering
\begin{tabular}{ccc}
\includegraphics[width=0.3\textwidth]{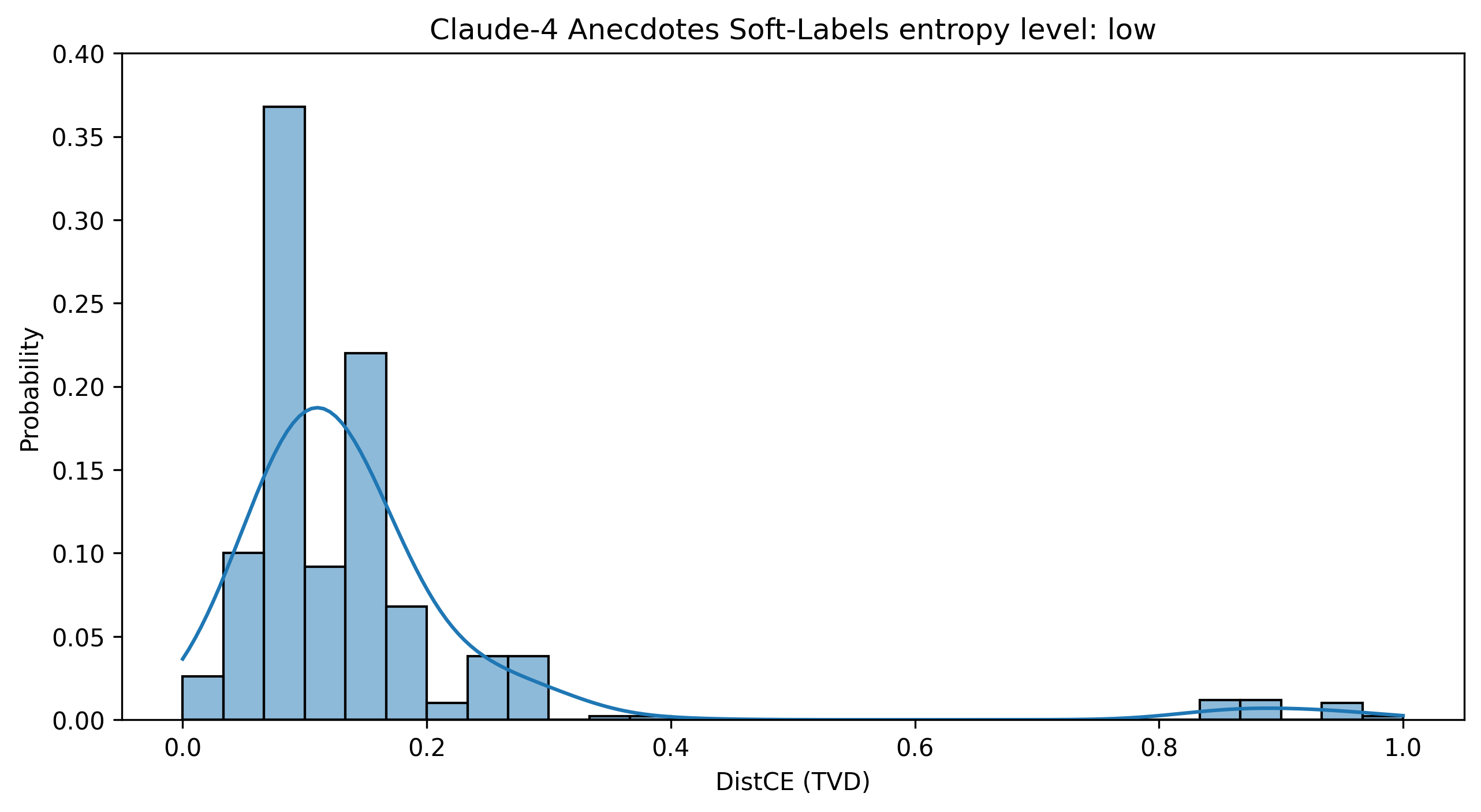} &
\includegraphics[width=0.3\textwidth]{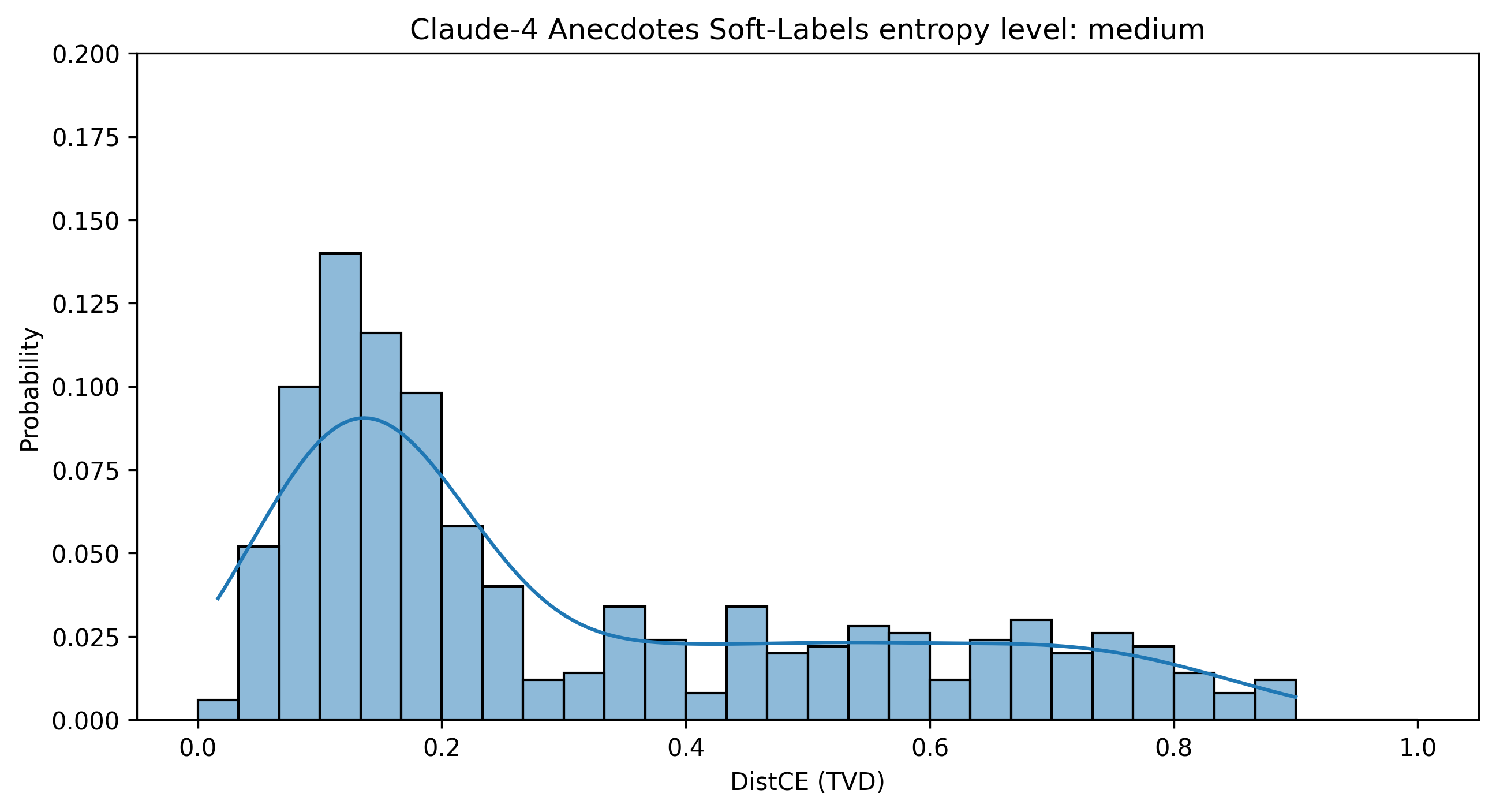} &
\includegraphics[width=0.3\textwidth]{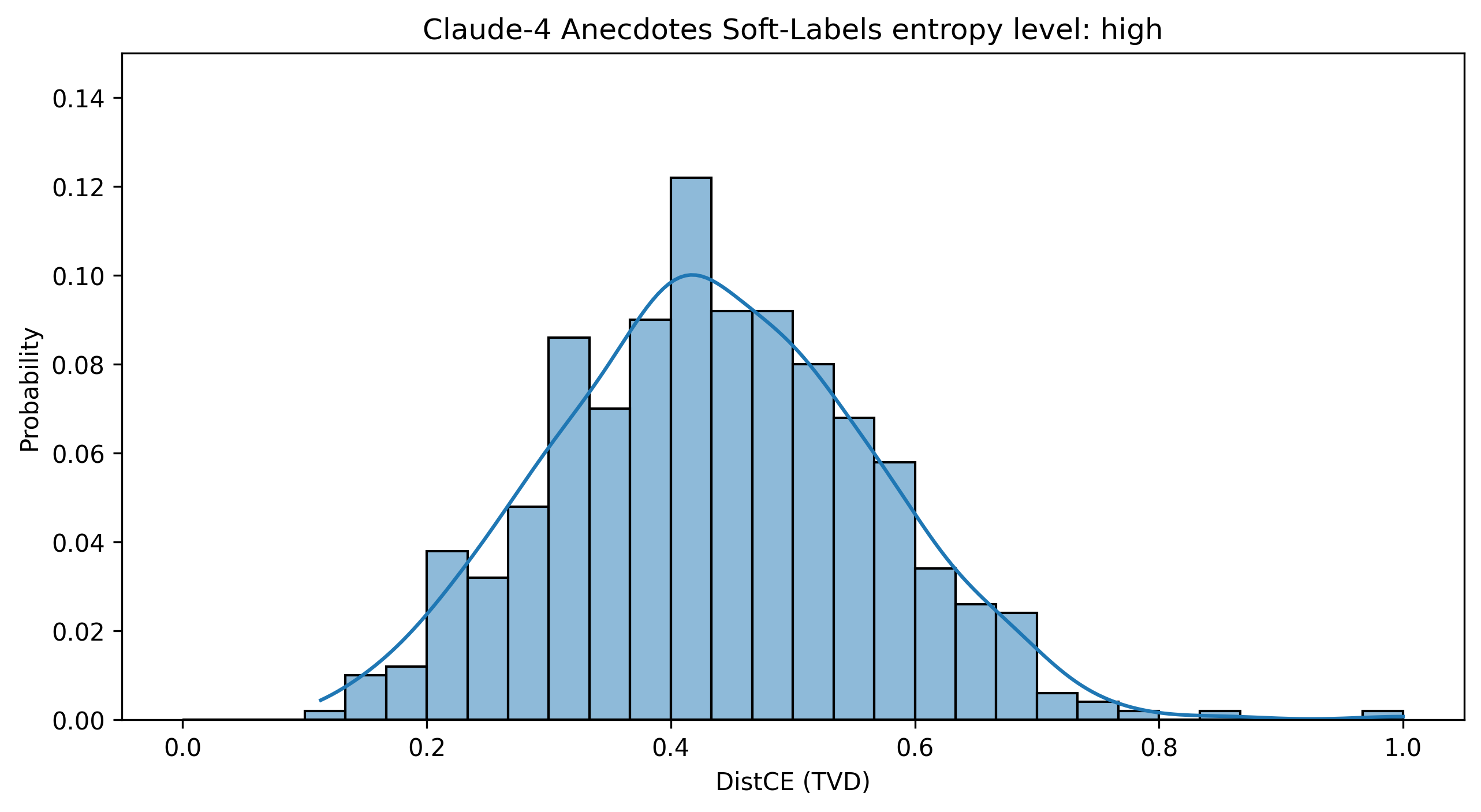} \\
(a) low $H$  & (b) medium $H$ & (c) high $H$  \\[1em]
\includegraphics[width=0.3\textwidth]{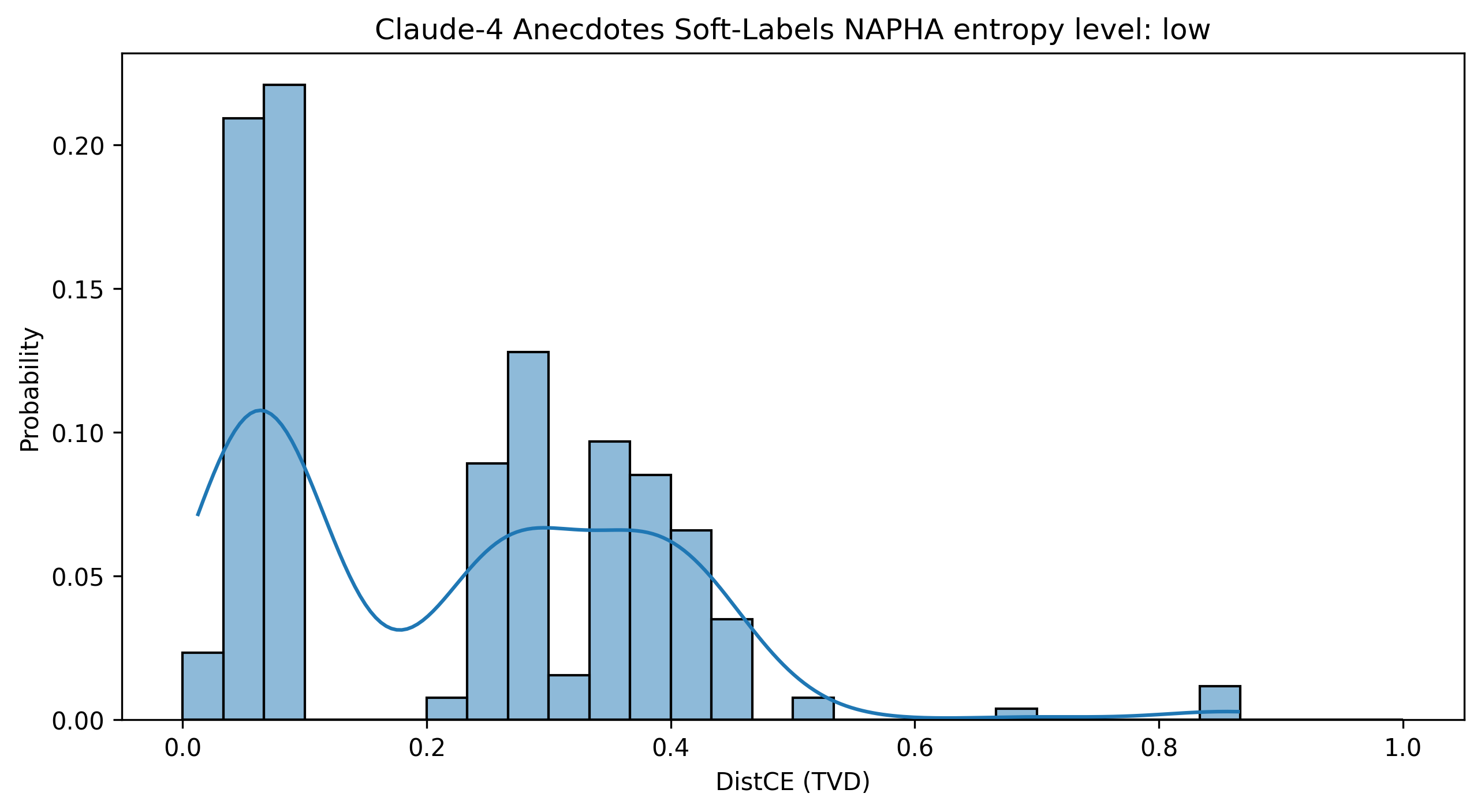} &
\includegraphics[width=0.3\textwidth]{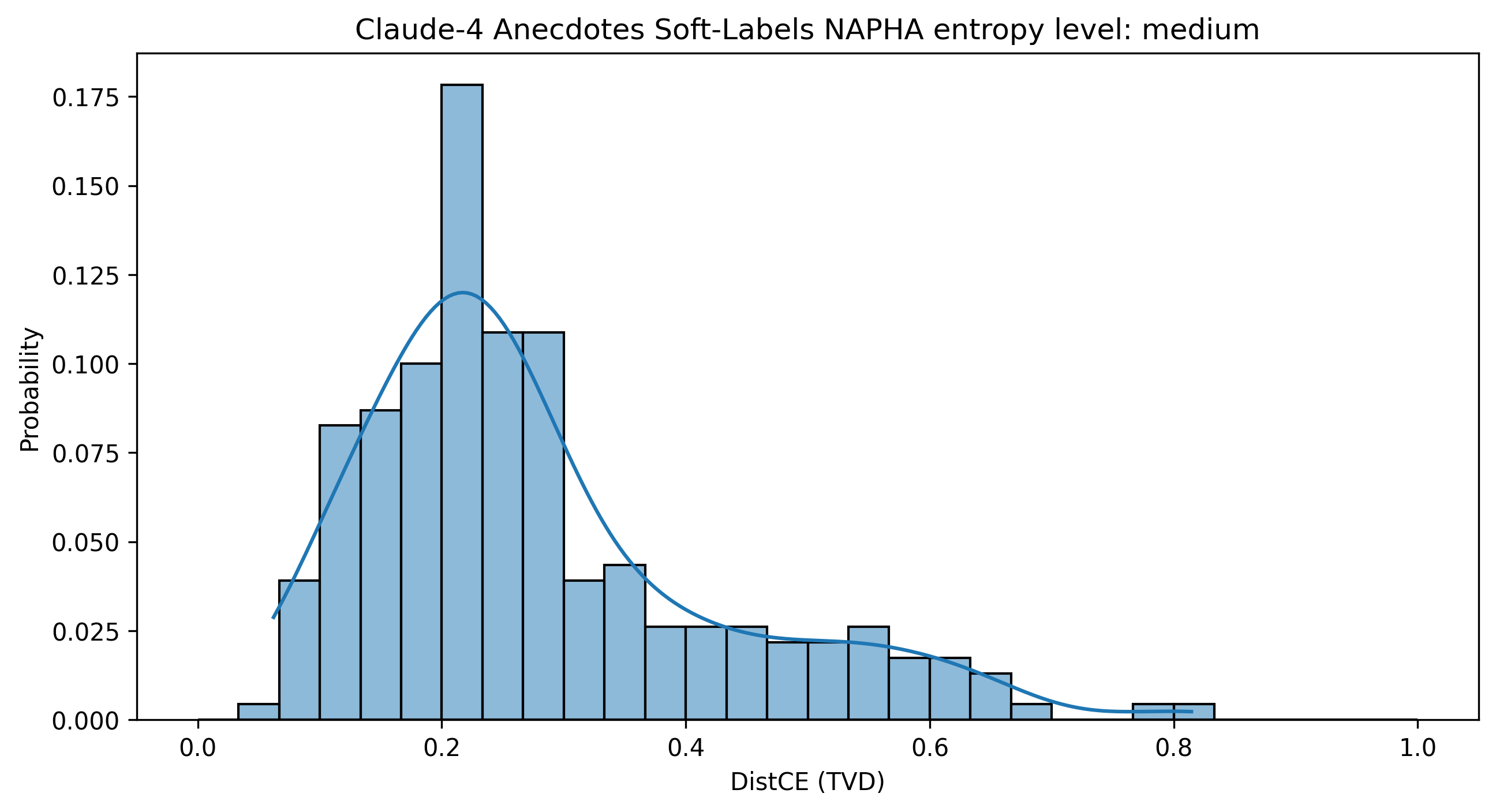} &
\includegraphics[width=0.3\textwidth]{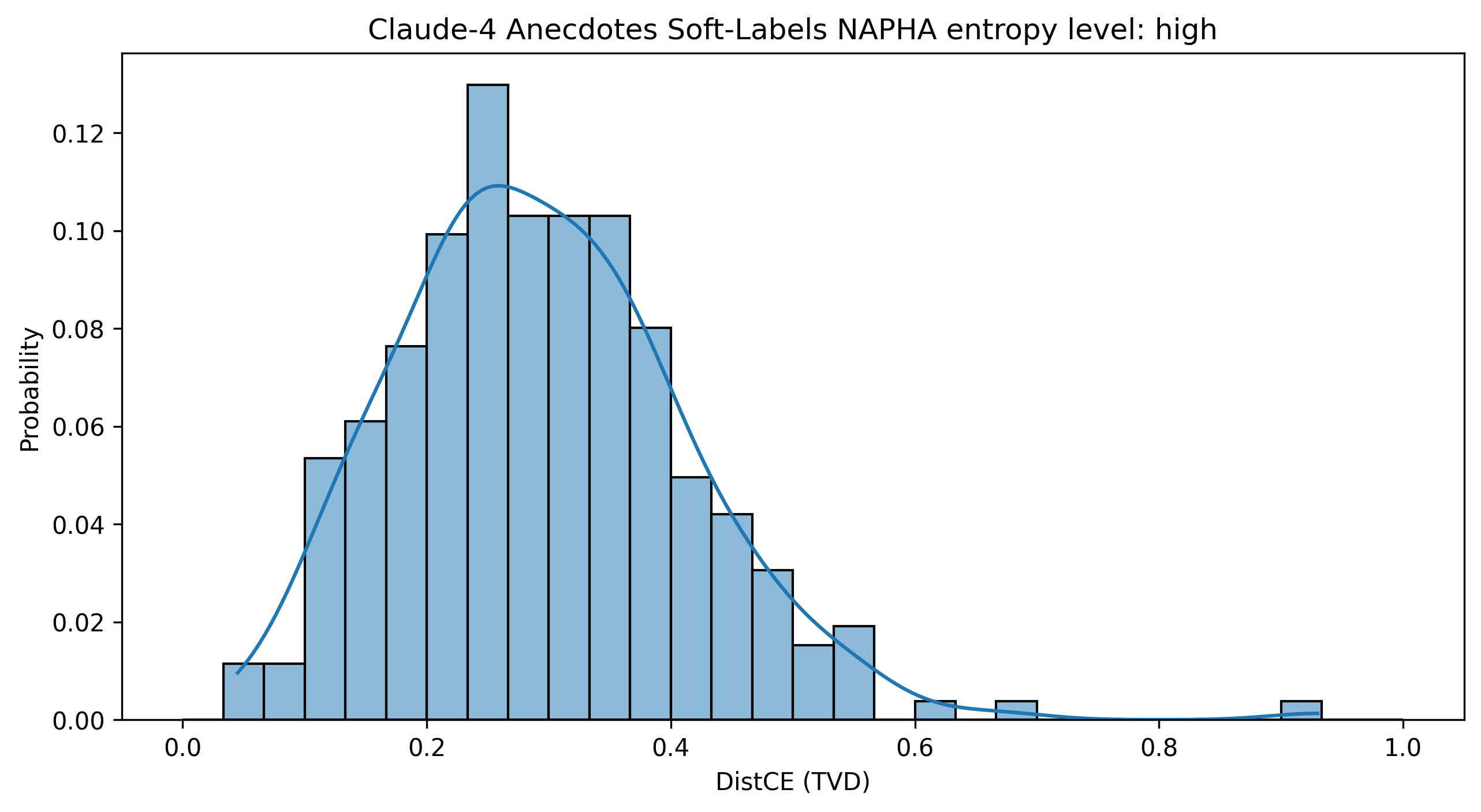} \\
(d) low $H$  & (e) medium $H$ & (f) high $H$  \\
\end{tabular}
\caption{Histogram plots of the DistCE metric on the Anecdotes dataset with 30 bins for low, medium and high entropy ($H$). The plots in the top row show Soft-Label prediction (hard ICL) without \methodName, and the ones at the bottom row after applying \methodName. We see that the probability mass shifts to the left, i.e. to better DistCE values, and tails get shorter, which is desirable.}
\label{fig:distCE_compare}
\end{figure*}

% After assessing that the  LLM-judges is able to perform similarly to human annotators on hard-labels, 

\subsubsection{Overall performance}
Table \ref{tab:result_soft_overview} shows \emph{soft-labels} prediction performance in the context of \ac{HLV} across different base models (SLP-HE, SLP-SE, SimAnn), with or without using \methodName.
We highlight that we observe small standard errors across our results, which suggests that our estimates are statistically robust.

First, we see that, \methodName improves  performance when used to calibrate predictions on top of base models.
This result is consistent across base models and datasets.
Indicatively, when we apply \methodName to SLP-SE, the DistCE metric  (the lower the better) decreases from 0.299 to 0.272 for the Anecdotes dataset, from 0.200 to 0.174 on ChaosNLI, and from 0.272 to 0.265 on DynaSent.
The JSD metric behaves in the same manner.
Importantly, we see large improvement in the oracle setting, which means using the oracle entropy class label for routing an instance to one of the three alignment models, instead of predicting the entropy class label.\footnote{We discuss the implications of this in the next section.}
In this setting, \methodName achieves a DistCE of 0.172 on Anecdotes, 0.153 on ChaosNLI, and 0.172 on DynaSent, all of which are substantial improvements over using the predicted entropy class labels.
As can be seen in Table~\ref{tab:ablation_otherCalibrationModels}, none of the other alignment models (Linear Transformation, Temperature Scaling, Dirichlet Calibration, and Parameterized Temperature Scaling) outperform ours.

We also calculate upper bound human performance on the datasets we consider as in~\citet{baan-etal-2022-stop}.
 We report the results in Table~\ref{tab:humanPerformance} in Appendix~\ref{sec:appendix_humanPerformance}, where we see that these datasets have different levels of complexity. 
 Contrasting Table~\ref{tab:humanPerformance} and Table \ref{tab:result_soft_overview}, we can conclude that \methodName approximates human performance on the Anecdotes and DynaSent datasets, but still lacks behind on the rest of the datasets, where there is a lot room for improvement in future work.

% \todo{add comment on oracle prediction}

% overall we see that LLMs perform poorly on soft-label prediction.
%
% This is in contrast to hard-label prediction, which we described in the previous subsection.

\subsubsection{Performance per entropy class}
\label{sec:performance-per-entropy-class}

\begin{table*}[t]
\centering
\resizebox{\linewidth}{!}{%
\begin{tabular}{lcccc|cccccc}
\hline
 &  &  & \multicolumn{2}{c|}{Average} & \multicolumn{2}{c}{Low $H$} & \multicolumn{2}{c}{Medium $H$} & \multicolumn{2}{c}{High $H$} \\
Base model & Alignment & Entropy Labels & DistCE $\downarrow$ & JSD $\downarrow$ & DistCE $\downarrow$ & JSD $\downarrow$ & DistCE $\downarrow$ & JSD $\downarrow$ & DistCE $\downarrow$ & JSD $\downarrow$ \\ \hline
{\multirow{3}{*}{SLP-HE}} & - & - & 0.299 & 0.326 & 0.155 & 0.250 & 0.308 & 0.350 & 0.435 & 0.366 \\
 & NAPHA & predicted & $0.271$ (0.001) & $0.292$ (0.001) & $0.228$ (0.004) & $0.299$ (0.003) & $0.295$ (0.002) & $0.313$ (0.002) & $0.292$ (0.004) & $0.263$ (0.003) \\
 & NAPHA & oracle & $0.175$ (0.002) & $0.227$ (0.002) & $0.040$ (0.004) & $0.129$ (0.005) & $0.274$ (0.002) & $0.305$ (0.002) & $0.210$ (0.001) & $0.212$ (0.001) \\
 \hline

\multirow{3}{*}{SLP-SE} & - & - & 0.299 & 0.326 & 0.196 & 0.279 & 0.311 & 0.352 & 0.388 & 0.341 \\
 & NAPHA & predicted & $0.272$ (0.002) & $0.294$ (0.001) & $0.226$ (0.004) & $0.295$ (0.003) & $0.290$ (0.002) & $0.316$ (0.002) & $0.299$ (0.003) & $0.269$ (0.002) \\
 & NAPHA & oracle  & $0.172$ (0.001) & $0.224$ (0.001) & $0.034$ (0.003) & $0.119$ (0.003) & $0.274$ (0.002) & $0.304$ (0.001) & $0.209$ (0.001) & $0.211$ (0.001) \\
 \hline

\multirow{3}{*}{SimAnn} & none & none & 0.325 & 0.385 & 0.053 & 0.163 & 0.375 & 0.413 & 0.550 & 0.497 \\
 & NAPHA & predicted & $0.293$ (0.001) & $0.308$ (0.001) & $0.271$ (0.006) & $0.321$ (0.004) & $0.277$ (0.002) & $0.313$ (0.001) & $0.330$ (0.005) & $0.289$ (0.004) \\
 & NAPHA & oracle & $0.172$ (0.001) & $0.225$ (0.001) & $0.035$ (0.003) & $0.120$ (0.005) & $0.273$ (0.002) & $0.304$ (0.001) & $0.209$ (0.001) & $0.211$ (0.001) \\

 \hline
\end{tabular}
} % end resizebox
\caption{Results on soft-label prediction per entropy class for the Anecdotes dataset. We report mean and standard error from 20 runs. Results on the per entropy class are shown in Appendix~\ref{sec:appdx_fullResults}.} 
\label{tab:results_soft_anecdotes}
\end{table*}

We show performance on each dataset per entropy class  in Tables \ref{tab:results_soft_anecdotes},  \ref{tab:results_soft_nli}, \ref{tab:results_soft_sentiment}, \ref{tab:results_soft_sme}, and \ref{tab:results_soft_topical}.
For the base model without \methodName, soft-label prediction worsens with increasing entropy (evidenced by DistCE and JSD).\footnote{
This is particularly prevalent for the SimAnn base model~\cite{simulatedAnnots}, which, due to its design and the relative stability of its predictions, performs much better on the low entropy class and much worse on high entropy cases w.r.t the other two base models.
}
% For the base model without \methodName, we see that the higher the entropy, the worse the base model becomes at predicting the soft-labels, as measured
% by the increase in DistCE and JSD metrics.
% While one might argue that is to be expected since high entropy can be interpreted as high uncertainty in labeling, which in turn indicates harder instances, this is not always true in the context of HLV.
% Here, a high entropy may indicate that multiple labels given by humans are valid under different perspectives, i.e., assumptions, values, and understandings.
% Put differently, each human annotator can be totally confident in their label choice, but the entropy on population level might still be high. 
% From this point of view, the high entropy class is the one where we desire the model to consider different opinions the most, and thus,  good alignment on the that class is particularly important. 
% Indeed, \methodName consistently improves performance on high entropy class instances.
While high entropy is often seen as labeling uncertainty, in HLV it can signal valid, diverse perspectives. Thus, even if each annotator is certain, population-level entropy can be high. Considering these perspectives is the most desirable, and thus, good alignment is particularly important.
Indeed, \methodName consistently improves performance on these high entropy instances.

While overall alignment improved with \methodName compared to just using the base models, alignment in the low entropy class became worse. % \todo{before we say performance is better at low entropy}
%which is evident when comparing Fig. \ref{fig:distCE_compare} (a) and (d), and also from Table \ref{tab:results_soft_anecdotes} in Appendix~\ref{sec:appdx_fullResults}.
%
This suggests that LLMs are by default better aligned on instances with low entropy soft-labels than on high entropy instances.
We attribute this effect to incorrect entropy label routing in \methodName, as this effect disappears when we use oracle entropy labels.
A simple but practical mitigation strategy for this downside of NAPHA is to not apply the method to the low entropy stratum.
In fact, \methodName can achieve very good alignment using oracle entropy classes, as seen in Figure \ref{fig:entropyAnalysis} in Appendix~\ref{sec:analysis-entropy}.
This shows that as entropy class assignment improves (which is arguably a simpler task than predicting the \emph{soft-labels} themselves), alignment to soft-labels using \methodName will be further improved.\footnote{
We find that LLMs do not perform well at predicting the oracle entropy class using standard prompting, as shown in Tab. \ref{tab:entropyClass_f1} in the Appendix. Note that it is unclear how hard the task of predicting the entropy class is for humans; studying this is an interesting direction for future work. }

\methodName improves performance on medium entropy instances relative to the base models. However, the improvement is less pronounced than for high entropy instances, despite medium entropy cases showing similar absolute distances to soft-label distributions.
This can be explained by the fact that the entropy is maximal for uniform distributions (high entropy class). 
%Thus, when assigning every possible label similar probability, the entropy is higher than when prioritizing only a subset of labels.
Thus, in cases where \emph{soft-labels}  are (close to) a uniform distribution, the straightforward solution of assigning equal likelihood to every entropy class will achieve good performance.
In contrast, predicting that an instance belongs to the medium entropy class is more challenging; this would require predicting that annotator votes are split between a subset of labels (e.g., 2/5 labels get a similar number of votes).

Finally, Figure \ref{fig:distCE_compare} further illustrates the behavior of the DistCE distance metric when applying \methodName on top of base models.
We observe that, when we apply \methodName, the probability mass shifts to the left, i.e. towards lower distances, which is the desired behavior.
This is especially evident in the high entropy class.
%
% In the low and medium entropy classes, we see that the tails became shorter, which is also desirable.
For the low and medium entropy classes, we observe shorter distribution tails, which indicates better alignment.

\subsection{Qualitative Analysis}
Since we use a chain-of-thought inducing statement in our prompts, we can analyze the model's reasoning traces qualitatively.
From this we found that the reasoning traces mostly do not consider alternative cultural perspectives, where the same behavior might be interpreted differently by humans. 
Also, we found that the model can default to a conflict-averse, mediating \textit{nobody is wrong} mode, whereas humans have polarized opinions. 
Appendix \ref{sec:appdx_caseStudy} shows and discusses two case-studies from the Anecdotes dataset.

\subsection{Ablations}
\label{sec:ablations}

\paragraph{Do separate alignment models per entropy class help?}
In Table \ref{tab:ablation_notEntropyAware} we show results of post-hoc alignment  when we only use a single alignment model instead of a separate alignment model per predicted entropy class as in \methodName.
Performance is comparable to using different alignment models per entropy class but deteriorates on low entropy cases.
Importantly, without separate classifiers per entropy class, we cannot achieve the potential improvements that
\methodName enables when routing would be improved (see Section~\ref{sec:performance-per-entropy-class}).

\paragraph{Does using oracle entropy class labels during training help?}
Since \methodName needs the oracle (ground truth) \emph{soft-labels} for training the alignment model, we can in principle use them instead of the predicted ones also during training time and only use the predicted ones at test time. However, since the prediction of entropy classes is inaccurate, using oracle entropy labels during training does not bring improvements (see Table \ref{tab:ablation_oraclePredicted}).

\paragraph{How much training data does \methodName need?}
Table \ref{tab:ablation_trainingData} shows the performance when using different amounts of training data for \methodName. We find that performance stabilizes at 10\% of training data, which shows that \methodName does not need a lot of human labels to improve alignment.

\section{Conclusion and Future Work}
This study demonstrates that \ac{LLMaJ} achieves close to human-level performance on \emph{hard-label} prediction tasks across  evaluated datasets.
However, our results show that the transition from \emph{hard-label} to \emph{soft-labels} prediction reveals significant limitations in current \ac{LLMaJ}, exposing its inability to adequately capture \ac{HLV}.
This is particularly relevant in scenarios where different human perspectives could lead to multiple labels being true simultaneously, most notably subjective tasks.
To address this, we proposed \methodName, a post-hoc alignment method for \emph{soft-labels} prediction that provides consistent improvements across all base models and datasets.
We observed substantial gains for high entropy instances, where capturing diverse human perspectives is most critical.
Furthermore, our oracle entropy class experiments revealed that as entropy class assignment improves, \methodName's effectiveness and practical applicability will  increase.
Also, \methodName can be extended to consider white-box access to models, e.g. by leveraging the distribution of logits across the possible labels.
More generally, future work should advance  \ac{HLV} approaches in two key areas. 
For model development, enabling \ac{HLV} might include fundamental changes to training regimes, where LLMs are rewarded the most for correctly representing pluralism.
For data collection, it is important to be able to disentangle human annotation error from valid \ac{HLV}.
To this end, beyond collecting labels from multiple human annotators, one should also collect explanations and confidence indications~\cite{chen-etal-2025-threading} as well as metadata such as sociodemographic information~\cite{sorensen-etal-2025-value}, which might be used to improve soft-label predictions. %~\cite{weber-genzel-etal-2024-varierr, sorensen-etal-2025-value,neumann2025usellmssimulateopinions,parappan-henao-2025-learning}.

\section*{Limitations}
While our study revealed interesting findings regarding the performance of \ac{LLMaJ} on predicting \emph{soft-labels}, and how to improve it with alignment methods, it is not without limitations.
First, the \emph{soft-labels} for three out of the five used datasets is sparse, due to low number of annotators (less than 6).
However, datasets with large number of annotators are rare, and expensive to collect, and thus using three to five annotators for human annotation is common practice.
It is unclear how many annotators are needed to get a fair estimate of the \emph{soft-labels}.
Our method uses the model's predicted label distribution for routing, instead of a separate uncertainty estimate. This may conflate model uncertainty with the predicted human disagreement, and cannot be disentangled with the current design.
Also, the way we sample the Anecdotes and DynaSent subsets is not according to their natural entropy-distribution. Instead, we explicitly construct a setting where the entropy classes are balanced, to get more detailed insights into the influence of entropy on the alignment. Note that for the three other datasets, we use them as-is, thus also preserving the natural entropy-distribution.
Moreover, since we calculate the entropy statistics across the training dataset, single instances can be misclassified. We can thus not ensure consistent improvement for every instance, but only show improvement across the entropy-stratum- and dataset-level.
Next, even when using open-weight models, we treat them as if we only had closed-weight access. We do not test the potential benefits from accessing model internals, which can be studied in future work.
Lastly, we did not perform hyperparameter tuning or tried to optimize the neural network architecture of our alignment models, focusing instead on showcasing the potential of \methodName with a lightweight approach.

%\section*{Acknowledgments}

\bibliography{custom}

\FloatBarrier
\appendix

\section{Theoretical Formalization of NAPHA}
\label{sec:formalization}

To establish the theoretical foundation of the NAPHA framework, we formalize the alignment of LLMs to the HJD as a partitioned optimization problem.

\textbf{Problem Definition and Objective.}
Let an instance be evaluated over $n$ possible discrete labels. The aggregated human annotations form a ground-truth probability distribution $y \in [0,1]^n$, where $\sum_{j=1}^n y_j = 1$. The base LLM outputs an initial soft-label prediction $\hat{y} \in [0,1]^n$. Our experiments found this to be poorly aligned with the ground-truth.

The goal of post-hoc alignment is to learn a mapping function $M_\phi$ (parameterized by weights $\phi$) that transforms the initial prediction $\hat{y}$ into a aligned distribution $\hat{y}_{cal} = M_\phi(\hat{y})$ that closely approximates $y$. 

To measure the distance between the target distribution $y$ and the aligned distribution $\hat{y}_{cal}$, we utilize the Kullback-Leibler (KL) divergence:
\begin{equation}
    D_{KL}(y \parallel \hat{y}_{cal}) = \sum_{j=1}^n y_j \log\left(\frac{y_j}{\hat{y}_{cal, j}}\right)
\end{equation}

\textbf{Entropy-Stratified Alignment.}
On the instance-level, the entropy of the human annotations will depend on different task-specific factors such as complexity, ambiguity, and different ethical understandings.
A single global alignment function $M_\phi$ may thus struggle to capture the diverse alignment patterns. To address this, NAPHA employs a piecewise approach by stratifying the data based on the Shannon entropy of the initial prediction:
\begin{equation}
    H(\hat{y}) = -\sum_{j=1}^n \hat{y}_j \log(\hat{y}_j)
\end{equation}
Based on $H(\hat{y})$, the input space is partitioned into three mutually exclusive subsets corresponding to low, medium, and high entropy instances (using terciles), denoted as $c \in \{L, M, H\}$. Instead of a single complex model, we learn a set of simpler, specialized alignment models $M_{\phi_c}$, such that the final alignment is defined as:
\begin{equation}
    \hat{y}_{cal} = \sum_{c \in \{L, M, H\}} \mathbb{I}(\hat{y} \in c) M_{\phi_c}(\hat{y})
\end{equation}
where $\mathbb{I}$ is the indicator function and can be understood as a router, routing the instances to their corresponding alignment model. By constraining the problem space for each $M_{\phi_c}$, we reduce the functional complexity required to achieve accurate alignment.

% This approach is theoretically motivated and inspired by error decomposition~\cite{breiman2001random, hastie2009elements}. We use a similar intuition to formulate NAPHA: by decomposing a highly complex global alignment problem into simpler, entropy-specific sub-problems, we aim to reduce the overall error.
This is analogous to mixture-of-experts architectures~\cite{jacobs1991adaptive}, where routing inputs to specialized models yields better overall performance than a single global model.

Let $\mathcal{E}(M)$ denote the expected alignment error of a mapping function $M$ over the entire data distribution. 
The expected global alignment error of our approach is the weighted sum of the conditional errors within each stratum $\mathcal{X}_c$:
\begin{align*}
    \mathcal{E}(\text{M}) =  \sum_{c \in \{L, M, H\}} P(x \in \mathcal{X}_c) \cdot \mathcal{E}(M_{\phi_c} \mid x \in \mathcal{X}_c),
\end{align*}

where $P(x \in \mathcal{X}_c)$ represents the probability of the data falling into entropy stratum $c$.

\section{Human Performance} \label{sec:appendix_humanPerformance}
We estimate the human performance by sampling 20\% annotations and comparing to the full distribution. We bootstrap 1000 runs. Table~\ref{tab:humanPerformance} shows the results. 
Due to the relatively low number of annotators on DynaSent, SummEval, and TopicalChat, the results have limited meaning on these datasets. 

\begin{table*}[tbh]
\centering
\resizebox{\linewidth}{!}{%
\begin{tabular}{lcc|cccccc}
\hline
 & \multicolumn{2}{c|}{Average} & \multicolumn{2}{c}{Low} & \multicolumn{2}{c}{Medium} & \multicolumn{2}{c}{High} \\
\multicolumn{1}{c}{Dataset} & DistCE $\downarrow$ & JSD $\downarrow$ & DistCE $\downarrow$ & JSD $\downarrow$ & DistCE $\downarrow$ & JSD $\downarrow$ & DistCE $\downarrow$ & JSD $\downarrow$ \\ \hline
ChaosNLI & 0.070 $\pm$ 0.024 & 0.095 $\pm$ 0.022 & 0.038 $\pm$ 0.013 & 0.091 $\pm$ 0.019 & 0.073 $\pm$ 0.027 & 0.093 $\pm$ 0.024 & 0.097 $\pm$ 0.039 & 0.105 $\pm$ 0.029 \\
Anecdotes & 0.165 $\pm$ 0.016 & 0.236 $\pm$ 0.016 & 0.014 $\pm$ 0.004 & 0.062 $\pm$ 0.005 & 0.1704 $\pm$ 0.020 & 0.231 $\pm$ 0.017 & 0.300 $\pm$ 0.029 & 0.332 $\pm$ 0.024 \\
DynaSent & 0.336 $\pm$ 0.057 & 0.399 $\pm$ 0.042 & 0 $\pm$ 0 & 0.007 $\pm$ 0 & 0.315 $\pm$ 0.148 & 0.359 $\pm$ 0.110 & 0.692 $\pm$ 0.025 & 0.584 $\pm$ 0.016 \\
SummEval & 0.264 $\pm$ 0.011 & 0.323 $\pm$ 0.010 & - & - & - & - & - & - \\
TopicalChat & 0.238 $\pm$ 0.009 & 0.318 $\pm$ 0.007 & - & - & - & - & - & - \\ \hline
\end{tabular}
} % resize box
\caption{Sampling 20\% (minimum one) human annotations and calculating DistCE and JSD metrics to the full distribution. Result shown as mean $\pm$ std from 1000 run bootstrapping.}
\label{tab:humanPerformance}
\end{table*}

\section{Analysis of HJD: sources of disagreement in the data}
\label{sec:analysis-of-hjd}
In this work, we perform experiments on datasets where HLV can have multiple sources besides annotation error. The datasets are chosen deliberately to cover a range of likely HLV sources. First, in the sentiment classification task (DynaSent), HLV might arise from ambiguity or sarcasm (see example in Fig. 1), which can be hard to detect in text. In SummEval and TopicalChat, HLV may more likely stem from individual preferences of, e.g., summarization style. The Anecdotes dataset on the other hand treats ethical dilemmas, and therefore can represent diverse, culturally-influenced, ethical understandings. Lastly, for ChaosNLI, the datapoints were chosen by the original dataset authors specifically for having elicited HLV in a previous study. Given that the task of NLI is mostly testing logical relations, which are objective, the HLV likely arises due to ambiguous or under-specified examples.
These sources of HLV can be conflated with our approach achieving human performance on the Anecdotes and DynaSent datasets, but lags behind it on the other datasets. For DynaSent and Anecdotes, disagreement stems from linguistic ambiguity and rich contextual information that LLMs are particularly good at interpreting, allowing NAPHA to better approximate human performance. For Anecdotes specifically, the long context input can already include hints on how the different parties in the story might be in the wrong. In contrast, ChaosNLI has consistently low distances across all entropy strata, setting an exceptionally high bar, while SummEval and TopicalChat involve inherently subjective evaluation criteria where classifying instances into entropy strata is much harder. It is precisely this subjectivity-driven HLV that LLMs struggle to capture. This finding is further supported by the oracle experiments on SummEval, which show that near-human performance is achievable if entropy classification were perfect, highlighting this as the key bottleneck.

\section{Results per entropy class }\label{sec:appdx_fullResults}
The full results that contain the per-entropy-class (or per-dimension) metrics are shown in Tables \ref{tab:results_soft_anecdotes} to \ref{tab:results_soft_topical}.

\begin{table*}[tbh]
\centering
\resizebox{\linewidth}{!}{%
\begin{tabular}{lcccc|cccccc}
\hline
 &  &  & \multicolumn{2}{c|}{Average} & \multicolumn{2}{c}{Low $H$} & \multicolumn{2}{c}{Medium $H$} & \multicolumn{2}{c}{High $H$} \\
Base model & Alignment & Entropy Labels & DistCE $\downarrow$ & JSD $\downarrow$ & DistCE $\downarrow$ & JSD $\downarrow$ & DistCE $\downarrow$ & JSD $\downarrow$ & DistCE $\downarrow$ & JSD $\downarrow$ \\ \hline
\multirow{3}{*}{SLP-HE} & none & none & 0.221 & 0.259 & 0.077 & 0.163 & 0.228 & 0.266 & 0.354 & 0.318 \\
 & NAPHA & predicted & $0.187$ (0.002) & $0.213$ (0.001) & $0.177$ (0.007) & $0.221$ (0.005) & $0.181$ (0.001) & $0.211$ (0.001) & $0.216$ (0.004) & $0.210$ (0.003) \\
 & NAPHA & oracle & $0.158$ (0.002) & $0.194$ (0.001) & $0.076$ (0.006) & $0.157$ (0.004) & $0.180$ (0.001) & $0.207$ (0.002) & $0.184$ (0.002) & $0.192$ (0.002) \\
 \hline
\multirow{3}{*}{SLP-SE} & none & none & 0.200 & 0.230 & 0.077 & 0.150 & 0.206 & 0.235 & 0.316 & 0.281 \\
 & NAPHA & predicted & $0.174$ (0.001) & $0.198$ (0.001) & $0.156$ (0.004) & $0.203$ (0.003) & $0.169$ (0.001) & $0.194$ (0.001) & $0.208$ (0.002) & $0.205$ (0.002) \\
 & NAPHA & oracle & $0.153$ (0.002) & $0.186$ (0.002) & $0.069$ (0.005) & $0.148$ (0.005) & $0.175$ (0.003) & $0.197$ (0.003) & $0.181$ (0.002) & $0.187$ (0.001) \\
\hline
\multirow{3}{*}{SimAnn} & none & none & 0.247 & 0.305 & 0.060 & 0.146 & 0.261 & 0.310 & 0.409 & 0.403 \\
 & NAPHA & predicted  & $0.190$ (0.002) & $0.216$ (0.001) & $0.184$ (0.005) & $0.220$ (0.004) & $0.176$ (0.002) & $0.209$ (0.002) & $0.238$ (0.003) & $0.228$ (0.002) \\
 & NAPHA & oracle & $0.144$ (0.001) & $0.181$ (0.001) & $0.043$ (0.001) & $0.107$ (0.001) & $0.170$ (0.001) & $0.199$ (0.001) & $0.181$ (0.001) & $0.189$ (0.001) \\
 \hline
\end{tabular}
} % end resizebox
\caption{Full results for soft-labels on ChaosNLI.}
\label{tab:results_soft_nli}
\end{table*}

\begin{table*}[tbh]
\centering
\resizebox{\linewidth}{!}{%
\begin{tabular}{lcccc|cccccc}
\hline
 &  &  & \multicolumn{2}{c|}{Average} & \multicolumn{2}{c}{Low $H$} & \multicolumn{2}{c}{Medium $H$} & \multicolumn{2}{c}{High $H$} \\ \hline
Base model & Alignment & Entropy Labels & DistCE $\downarrow$ & JSD $\downarrow$ & DistCE $\downarrow$ & JSD $\downarrow$ & DistCE $\downarrow$ & JSD $\downarrow$ & DistCE $\downarrow$ & JSD $\downarrow$ \\ \hline
\multirow{3}{*}{SLP-HE} & none & none & 0.268 & 0.313 & 0.079 & 0.181 & 0.253 & 0.324 & 0.474 & 0.394 \\
 & NAPHA & predicted & $0.263$ (0.003) & $0.306$ (0.002) & $0.228$ (0.009) & $0.306$ (0.005) & $0.293$ (0.002) & $0.344$ (0.002) & $0.269$ (0.004) & $0.259$ (0.002) \\
& NAPHA & oracle &  $0.158$ (0.002) & $0.236$ (0.002) & $0.043$ (0.004) & $0.143$ (0.005) & $0.248$ (0.005) & $0.324$ (0.002) & $0.182$ (0.002) & $0.204$ (0.001) \\
 \hline

\multirow{3}{*}{SLP-SE} & none & none & 0.272 & 0.330 & 0.116 & 0.219 & 0.238 & 0.321 & 0.463 & 0.419 \\
  & NAPHA & predicted & $0.265$ (0.002) & $0.305$ (0.001) & $0.217$ (0.007) & $0.298$ (0.005) & $0.285$ (0.003) & $0.339$ (0.002) & $0.293$ (0.005) & $0.274$ (0.003) \\
 & NAPHA & oracle & $0.172$ (0.004) & $0.249$ (0.003) & $0.083$ (0.011) & $0.193$ (0.009) & $0.252$ (0.004) & $0.325$ (0.002) & $0.182$ (0.002) & $0.204$ (0.001) \\
 \hline

\multirow{3}{*}{SimAnn} & none & none & 0.296 & 0.367 & 0.033 & 0.129 & 0.279 & 0.351 & 0.575 & 0.515 \\
 & NAPHA & predicted & $0.294$ (0.002) & $0.323$ (0.001) & $0.262$ (0.006) & $0.323$ (0.004) & $0.270$ (0.002) & $0.336$ (0.001) & $0.351$ (0.003) & $0.307$ (0.002)  \\
 & NAPHA & oracle & $0.158$ (0.004) & $0.238$ (0.003) & $0.052$ (0.010) & $0.152$ (0.010) & $0.241$ (0.005) & $0.322$ (0.002) & $0.180$ (0.001) & $0.204$ (0.001) \\
\hline
\end{tabular}
} % end resizebox
\caption{Full results for soft-labels on  DynaSent.}
\label{tab:results_soft_sentiment}
\end{table*}

\begin{table*}[tbh]
\centering
\resizebox{\linewidth}{!}{%
\begin{tabular}{lcccccccccc|cc}
\hline
 &  &  & \multicolumn{2}{c}{Relevance} & \multicolumn{2}{c}{Coherence} & \multicolumn{2}{c}{Consistency} & \multicolumn{2}{c|}{Fluency} & \multicolumn{2}{c}{Average} \\
Base model & Alignment & Entropy Labels & DistCE $\downarrow$ & JSD $\downarrow$ & DistCE $\downarrow$ & JSD $\downarrow$ & DistCE $\downarrow$ & JSD $\downarrow$ & DistCE $\downarrow$ & JSD $\downarrow$ & DistCE $\downarrow$ & JSD $\downarrow$ \\ \hline
\multirow{3}{*}{SLP-HE} & none & none & 0.503 & 0.488 & 0.516 & 0.503 & 0.218 & 0.349 & 0.494 & 0.520 & 0.433 & 0.465 \\
 & NAPHA & predicted & $0.409$ (0.001) & $0.421$ (0.001)  & $0.453$ (0.001) & $0.452$ (0.001) & $0.148$ (0.002) & $0.260$ (0.002)  &$0.202$ (0.002) & $0.291$ (0.001)& 0.303 & 0.356 \\
 & NAPHA & oracle & $0.386$ (0.003) & $0.415$ (0.002) & $0.421$ (0.002) & $0.441$ (0.001) & $0.110$ (0.003) & $0.232$ (0.003) & $0.133$ (0.002) & $0.249$ (0.002) & 0.262 & 0.334 \\ 
 \hline
\multirow{3}{*}{SLP-SE} & none & none & 0.462 & 0.501 & 0.470 & 0.512 & 0.172 & 0.319 & 0.381 & 0.482 & 0.371 & 0.453  \\
 & NAPHA & predicted & $0.405$ (0.001) & $0.418$ (0.001) & $0.445$ (0.001) & $0.447$ (0.001) & $0.152$ (0.003) & $0.266$ (0.002) & $0.210$ (0.002) & $0.299$ (0.001)  & 0.303& 0.358 \\
 & NAPHA & oracle & $0.386$ (0.002) & $0.414$ (0.001) &  $0.416$ (0.002) & $0.438$ (0.001) & $0.113$ (0.003) & $0.236$ (0.003) & $0.137$ (0.002) & $0.254$ (0.001)  & 0.263 & 0.335 \\ 
 \hline
\multirow{3}{*}{SimAnn} & none & none & 0.697 & 0.643 & 0.552 & 0.546 & 0.176 & 0.319 & 0.822 & 0.732 & 0.562 & 0.560  \\
 & NAPHA & predicted &$0.411$ (0.001) & $0.424$ (0.001) & $0.445$ (0.002) & $0.448$ (0.001) &$0.154$ (0.002) & $0.267$ (0.002)  & $0.210$ (0.002) & $0.300$ (0.001) & 0.305 & 0.360 \\
 & NAPHA & oracle & $0.389$ (0.002) & $0.417$ (0.001) &  $0.404$ (0.002) & $0.431$ (0.001) &  $0.107$ (0.002) & $0.229$ (0.002) & $0.132$ (0.002) & $0.250$ (0.001) & 0.258 & 0.332 \\ \hline
\end{tabular}
} % end resizebox
\caption{Full results for soft-labels on SummEval.}
\label{tab:results_soft_sme}
\end{table*}

\begin{table*}[tbh]
\centering
\resizebox{\linewidth}{!}{%
\begin{tabular}{lcccccccccc|cc}
\hline
 &  &  & \multicolumn{2}{c}{Naturalness} & \multicolumn{2}{c}{Coherence} & \multicolumn{2}{c}{Engagingness} & \multicolumn{2}{c|}{Groundedness} & \multicolumn{2}{c}{$\mu \pm \sigma$} \\
Base model & Alignment & Entropy Labels & DistCE $\downarrow$ & JSD $\downarrow$ & DistCE $\downarrow$ & JSD $\downarrow$ & DistCE $\downarrow$ & JSD $\downarrow$ & DistCE $\downarrow$ & JSD $\downarrow$ & DistCE $\downarrow$ & JSD $\downarrow$ \\ \hline
\multirow{3}{*}{SLP-HE} & none & none & 0.383 & 0.407 & 0.366 & 0.390 & 0.363 & 0.381 & 0.341 & 0.439 & 0.363  & 0.404 \\
 & NAPHA & predicted & $0.343$ (0.003) & $0.373$ (0.002) &  $0.344$ (0.003) & $0.373$ (0.003)  & $0.342$ (0.002) & $0.365$ (0.002) & $0.367$ (0.005) & $0.394$ (0.003) & 0.349 & 0.376 \\
 & NAPHA & oracle & $0.283$ (0.003) & $0.351$ (0.002) & $0.271$ (0.004) & $0.339$ (0.002) &$0.297$ (0.005) & $0.349$ (0.003) & $0.354$ (0.003) & $0.384$ (0.001) & 0.301 & 0.356 \\
 \hline
\multirow{3}{*}{SLP-SE} & none & none & 0.409 & 0.434 & 0.376 & 0.410 & 0.352 & 0.388 & 0.357 & 0.464 & 0.374  & 0.424  \\
 & NAPHA & predicted &$0.358$ (0.002) & $0.384$ (0.002) &$0.355$ (0.004) & $0.381$ (0.003)   & $0.349$ (0.002) & $0.374$ (0.002) &  $0.390$ (0.004) & $0.401$ (0.002) & 0.363 & 0.385 \\
 & NAPHA & oracle & $0.301$ (0.003) & $0.363$ (0.002) & $0.269$ (0.003) & $0.338$ (0.002) & $0.304$ (0.005) & $0.355$ (0.003)&$0.374$ (0.003) & $0.395$ (0.002) & 0.312 & 0.363 \\ 
 \hline
\multirow{3}{*}{SimAnn} & none & none & 0.413 & 0.456 & 0.353 & 0.413 & 0.369 & 0.412 & 0.359 & 0.449 & 0.374  & 0.433 \\
 & NAPHA & predicted &$0.339$ (0.003) & $0.369$ (0.002)& $0.343$ (0.003) & $0.372$ (0.002) &$0.326$ (0.003) & $0.354$ (0.002) &  $0.386$ (0.002) & $0.404$ (0.002) &0.349 &0.375 \\
 & NAPHA & oracle & $0.263$ (0.005) & $0.336$ (0.003)  & $0.245$ (0.003) & $0.321$ (0.002) & $0.256$ (0.005) & $0.318$ (0.003) & $0.364$ (0.004) & $0.391$ (0.003) & 0.282 & 0.342 \\ \hline
\end{tabular}
} % end resizebox
\caption{Full results for soft-labels on TopicalChat}
\label{tab:results_soft_topical}
\end{table*}

\section{Analysis of Entropy}
\label{sec:analysis-entropy}
\begin{figure*}[t]
\centering
\begin{tabular}{cc}
\includegraphics[width=0.45\textwidth]{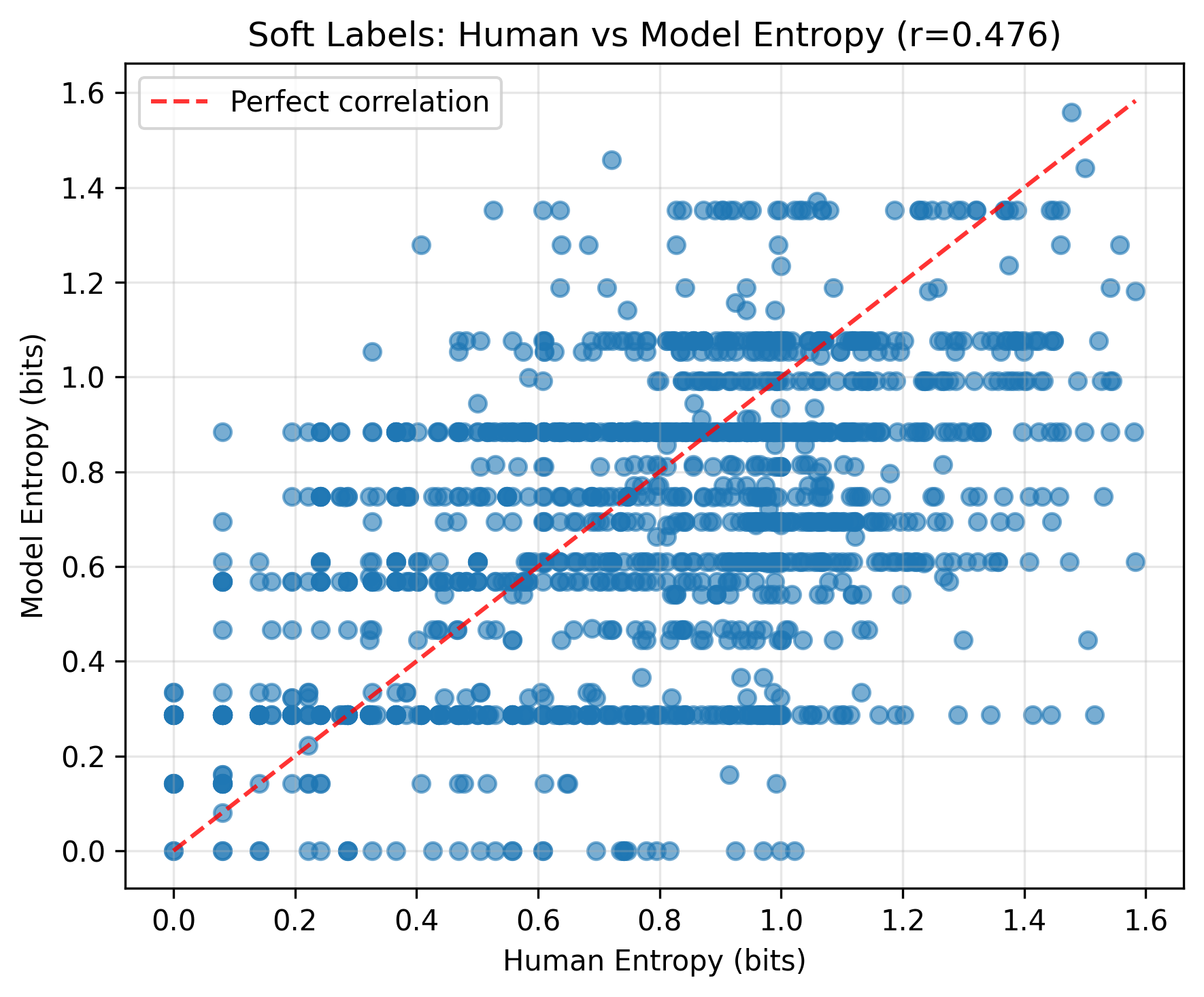} &
\includegraphics[width=0.45\textwidth]{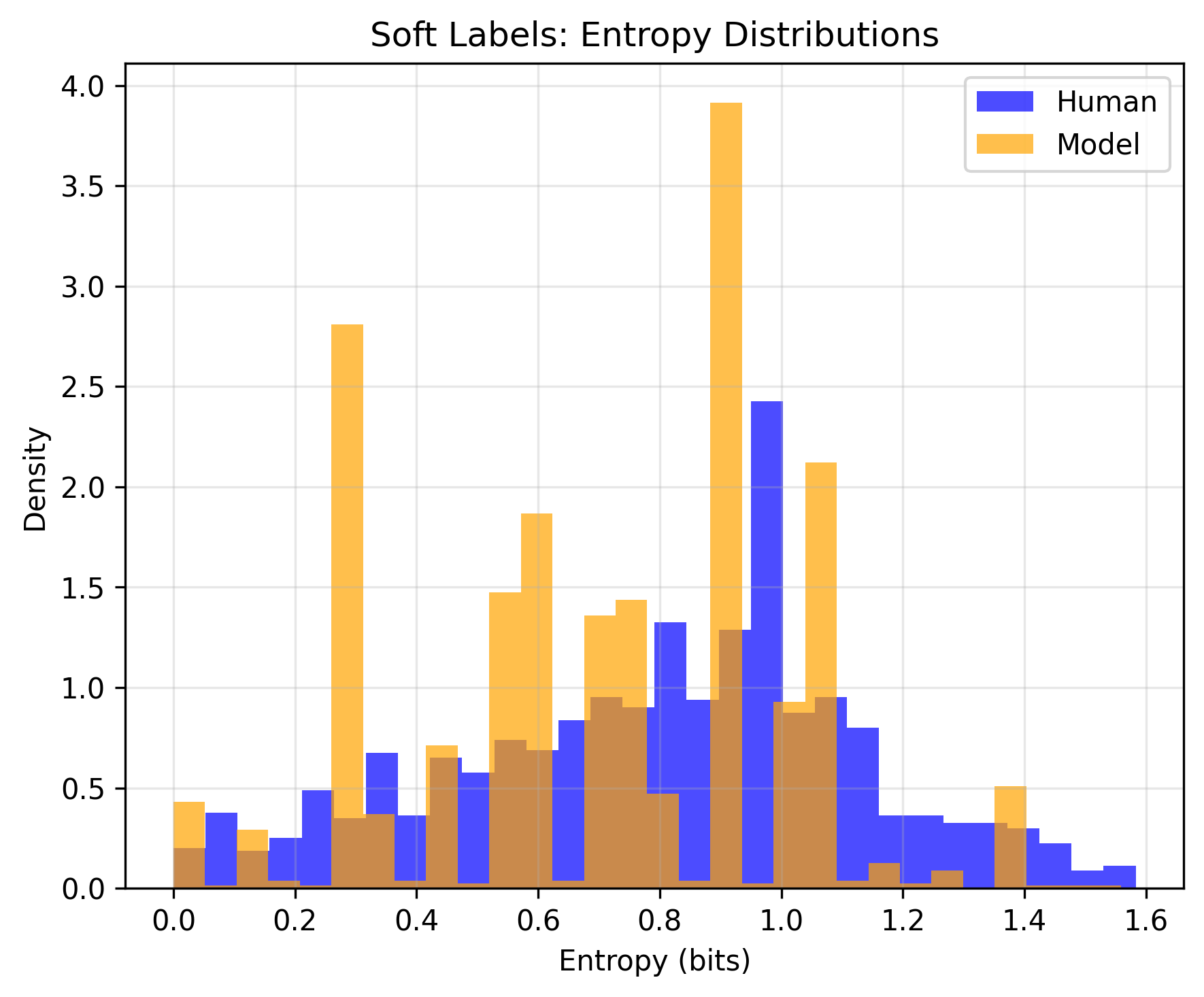} \\
(a) Correlation of human and model entropy & (b) Entropy Distribution \\[1em]

\includegraphics[width=0.45\textwidth]{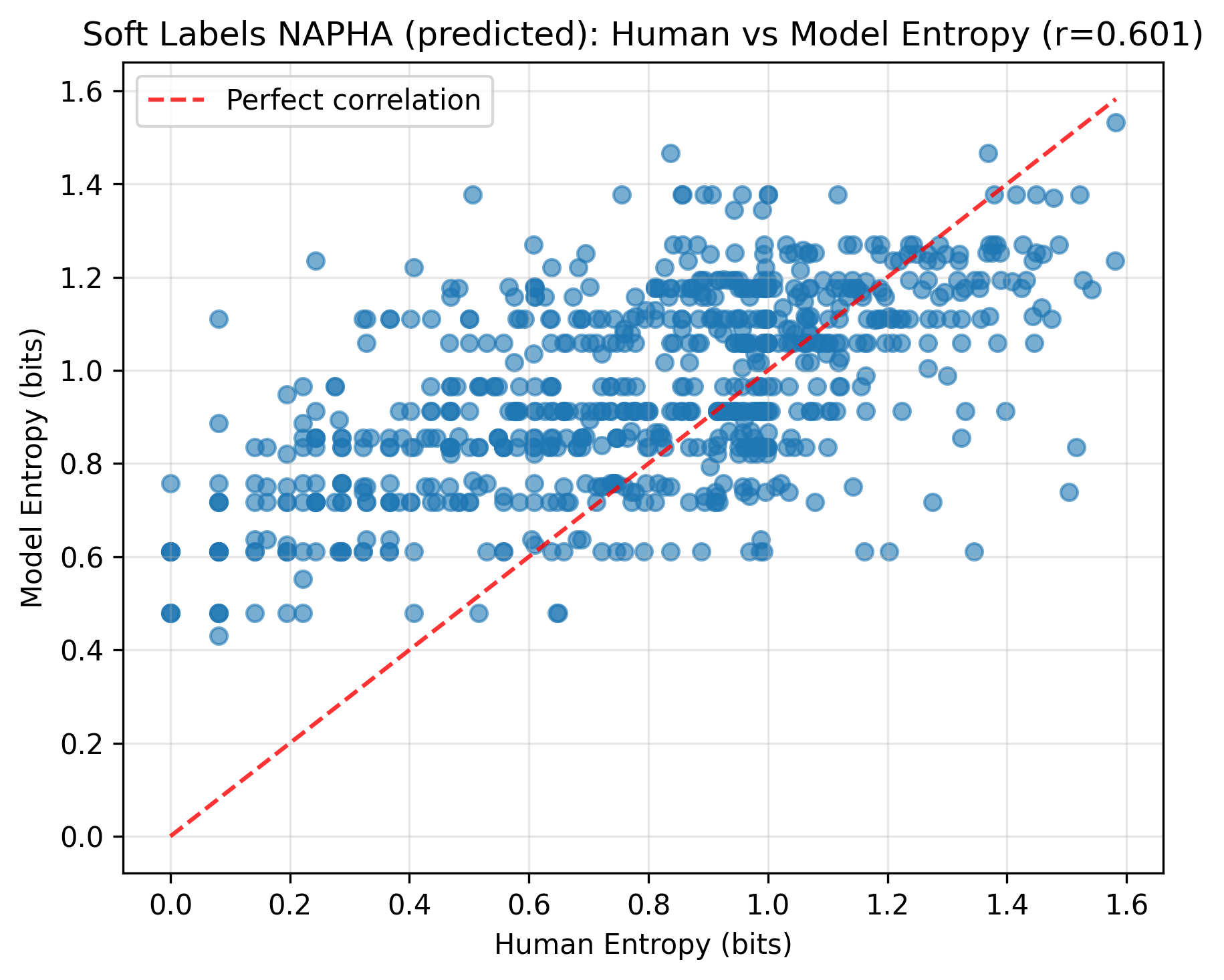} &
\includegraphics[width=0.45\textwidth]{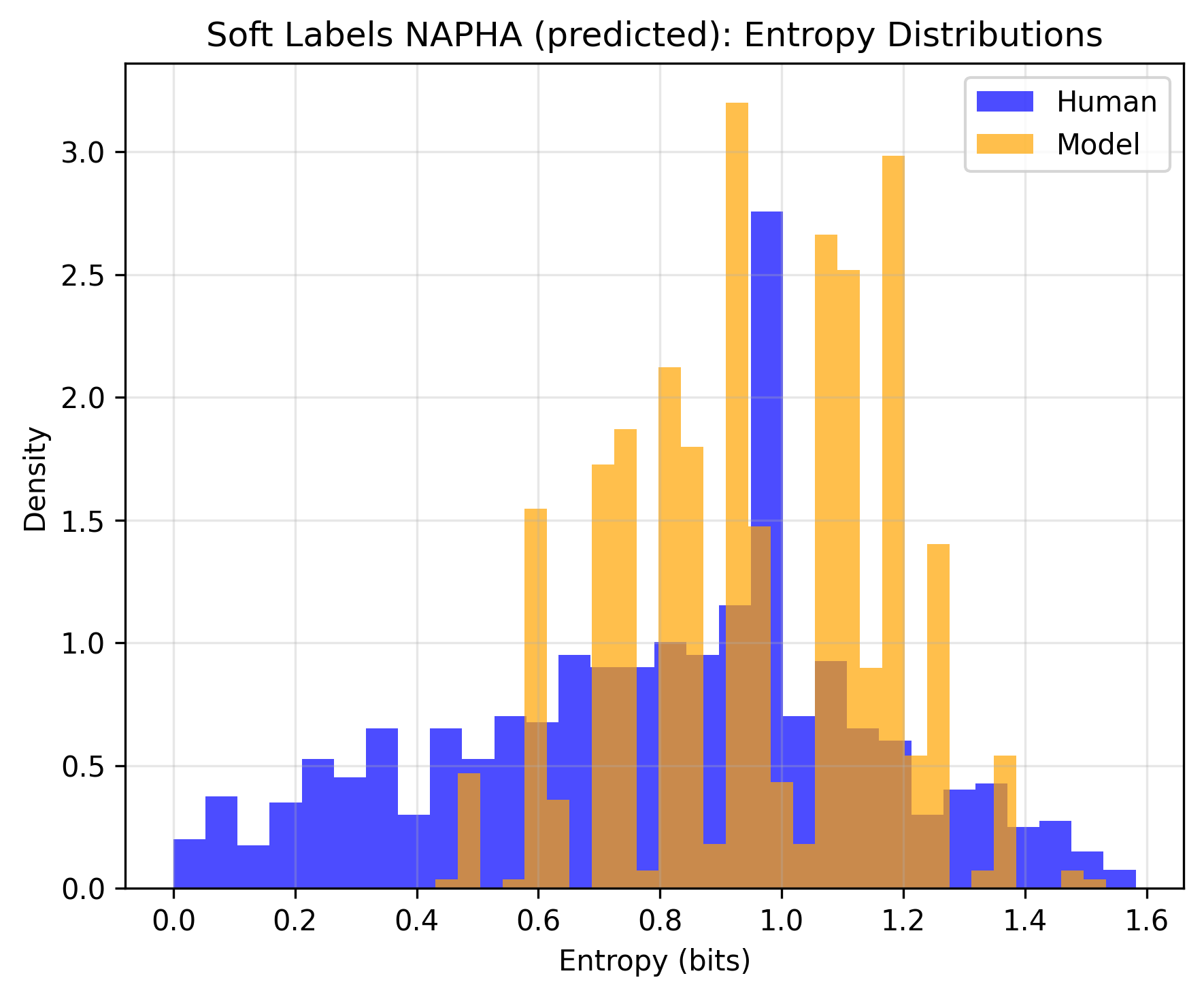} \\
(d) Correlation of human and model entropy & (d) Entropy Distribution \\[1em]

\includegraphics[width=0.45\textwidth]{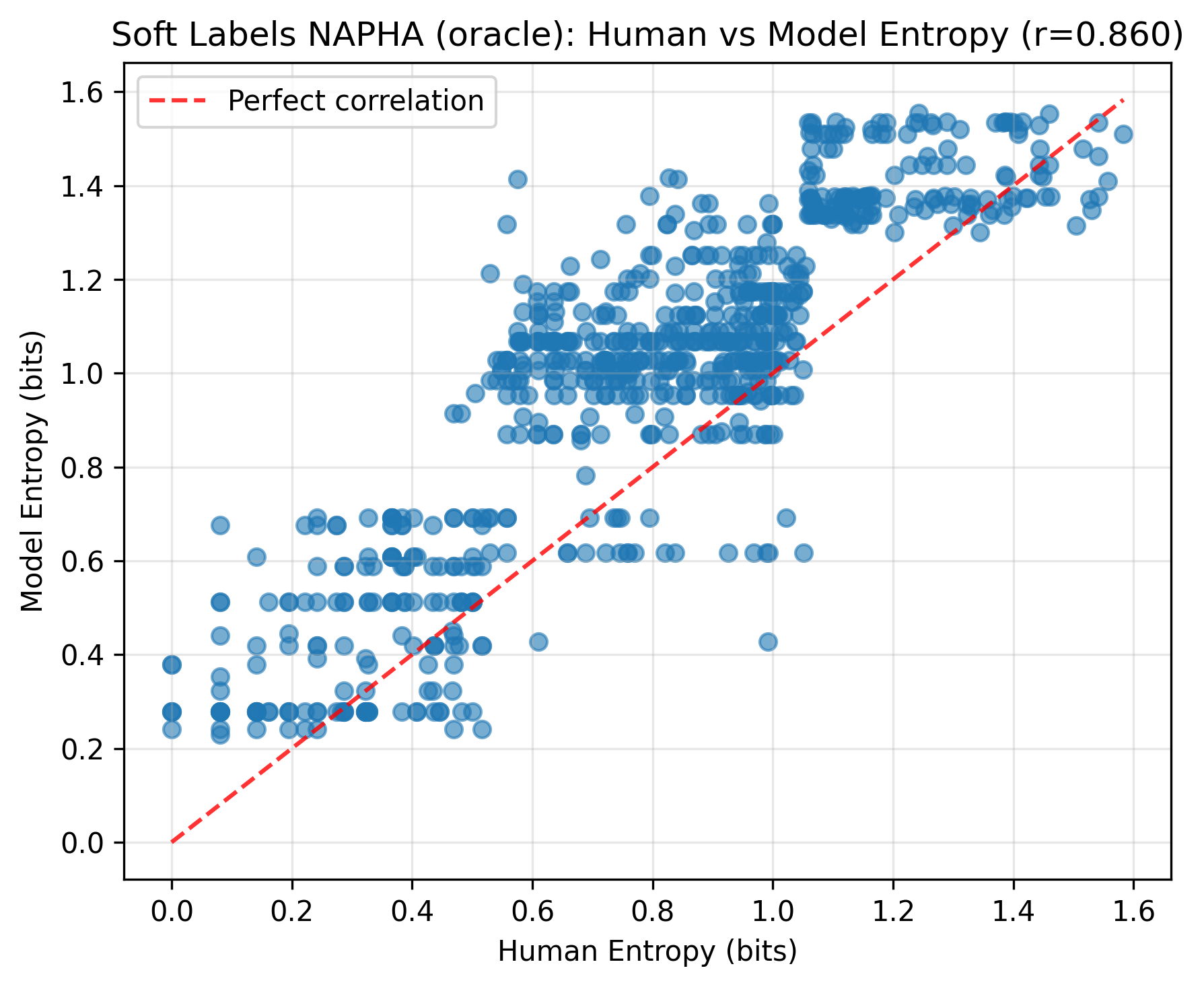} &
\includegraphics[width=0.45\textwidth]{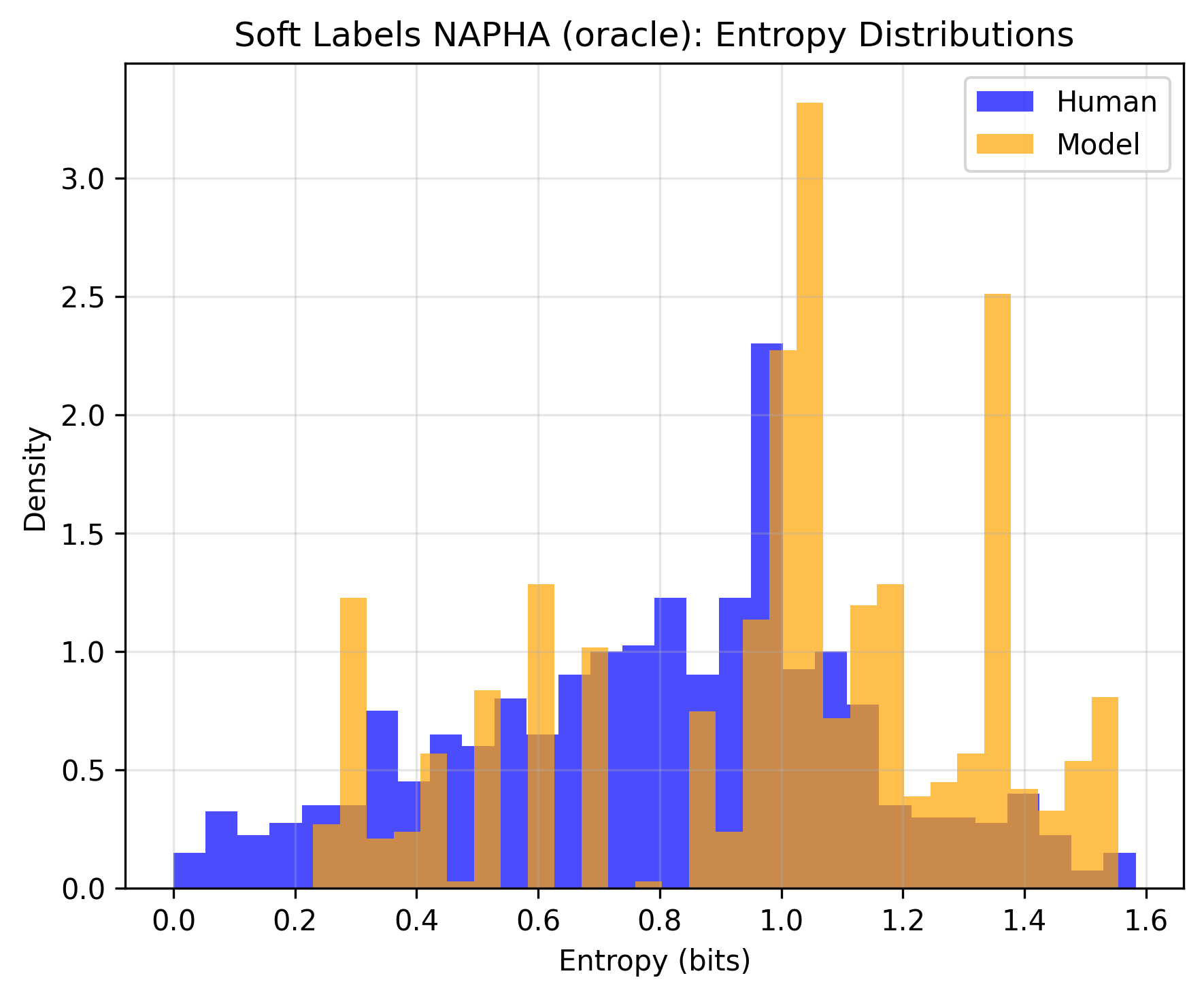} \\
(e) Correlation of human and model entropy & (f) Entropy Distribution \\
\end{tabular}
\caption{Analysis of entropy across the predictions. Top: Without NAPHA. Middle: With NAPHA with predicted entropy classes. Bottom: With NAPHA with oracle entropy classes.}
\label{fig:entropyAnalysis}
\end{figure*}

Fig. \ref{fig:entropyAnalysis} illustrates an analysis of the entropy across the predictions with and without \methodName. Table \ref{tab:entropyClass_f1} shows the F1-scores of predicting the entropy class.

\begin{table}[]
\centering
\resizebox{\linewidth}{!}{%
\begin{tabular}{lccc}
\hline
Dataset & \begin{tabular}[c]{@{}c@{}}Soft-Labels \\ hard ICL\end{tabular} & \begin{tabular}[c]{@{}c@{}}Soft-Labels\\ soft ICL\end{tabular} & \begin{tabular}[c]{@{}c@{}}Simulated \\ Annotators\end{tabular} \\ \hline
ChaosNLI & 0.53 & 0.56 & 0.39 \\
Anecdotes & 0.51 & 0.46 & 0.50 \\
DynaSent &  0.52 & 0.53 & 0.48 \\
SummEval & 0.46 & 0.72 & 0.49 \\
TopicalChat & 0.35 & 0.54 & 0.50 \\ \hline
\end{tabular}
} % end resizebox
\caption{Weighted Average F1 Score for the classification of the entropy class.}
\label{tab:entropyClass_f1}
\end{table}

\section{Ablations} \label{sec:appx_ablations}
% The following contains different ablation studies. 
First, Tab. \ref{tab:ablation_notEntropyAware} shows the performance of post-hoc alignment without being entropy-aware.
Next, Tab. \ref{tab:ablation_oraclePredicted} shows the performance when using oracle entropy-classes during training of the \methodName alignment models, but predicted entropy-classes during test-set inference.
In Tab. \ref{tab:ablation_trainingData} we include results when using different test-train splits. 
Finally, Tab. \ref{tab:ablation_otherCalibrationModels} showcases results when using different architectures for the alignment models. 
While the linear transformation and temperature scaling \cite{guo2017calibration} show slightly worse performance, the Dirichlet calibration \cite{kull2019beyond} presents with virtually the same performance. Thus, we consider both using a neural network and Dirichlet calibration as equally valid choices for the post-hoc alignment model architecture.

\begin{table*}[]
\centering
\resizebox{\linewidth}{!}{%
\begin{tabular}{lcc|cccccc}
\hline
 & \multicolumn{2}{c|}{Average} & \multicolumn{2}{c}{Low} & \multicolumn{2}{c}{Medium} & \multicolumn{2}{c}{High} \\
\multicolumn{1}{c}{Dataset} & DistCE $\downarrow$ & JSD $\downarrow$ & DistCE $\downarrow$ & JSD $\downarrow$ & DistCE $\downarrow$ & JSD $\downarrow$ & DistCE $\downarrow$ & JSD $\downarrow$ \\ \hline
ChaosNLI & 0.185 & 0.203 & 0.204 & 0.235 & 0.169 & 0.189 & 0.209 & 0.205 \\
Anecdotes & 0.268 & 0.288 & 0.216 & 0.285 & 0.294 & 0.313 & 0.294 & 0.262 \\
DynaSent & 0.278 & 0.310 & 0.252 & 0.319 & 0.278 & 0.336 & 0.306 & 0.271 \\
SummEval & 0.297 & 0.352 & - & - & - & - & - & - \\
TopicalChat & 0.355 & 0.376 & - & - & - & - & - & - \\ \hline
\end{tabular}
}
\caption{Using Post-hoc alignment without being entropy-aware. Thus, only one alignment model is used for all datapoints. Results shown for base model SLP-SE.}
\label{tab:ablation_notEntropyAware}
\end{table*}

\begin{table*}[]
\centering
\resizebox{\linewidth}{!}{%
\begin{tabular}{lcc|cccccc}
\hline
 & \multicolumn{2}{c|}{Average} & \multicolumn{2}{c}{Low} & \multicolumn{2}{c}{Medium} & \multicolumn{2}{c}{High} \\
\multicolumn{1}{c}{Dataset} & DistCE $\downarrow$ & JSD $\downarrow$ & DistCE $\downarrow$ & JSD $\downarrow$ & DistCE $\downarrow$ & JSD $\downarrow$ & DistCE $\downarrow$ & JSD $\downarrow$ \\ \hline
ChaosNLI & 0.189 & 0.219 & 0.130 & 0.191 & 0.202 & 0.228 & 0.218 & 0.220 \\
Anecdotes & 0.289 & 0.306 & 0.285 & 0.332 & 0.295 & 0.315 & 0.287 & 0.266 \\
DynaSent & 0.269 & 0.308 & 0.196 & 0.281 & 0.275 & 0.335 & 0.336 & 0.305 \\
SummEval & 0.320 & 0.371 & - & - & - & - & - & - \\
TopicalChat & 0.383 & 0.401 & - & - & - & - & - & - \\ \hline
\end{tabular}
}
\caption{Using oracle entropy labels for training, and predicted labels for testing. Results shown for base model SLP-SE.}
\label{tab:ablation_oraclePredicted}
\end{table*}

\begin{table*}[]
\centering
\resizebox{\linewidth}{!}{%
\begin{tabular}{lccc|cccccc}
\hline
 &  & \multicolumn{2}{c|}{Average} & \multicolumn{2}{c}{Low} & \multicolumn{2}{c}{Medium} & \multicolumn{2}{c}{High} \\
Dataset & Training \% & DistCE $\downarrow$ & JSD $\downarrow$ & DistCE $\downarrow$ & JSD $\downarrow$ & DistCE $\downarrow$ & JSD $\downarrow$ & DistCE $\downarrow$ & JSD $\downarrow$ \\ \hline
\multirow{4}{*}{ChaosNLI} & 50\% & 0.171 & 0.197 & 0.140 & 0.188 & 0.171 & 0.196 & 0.206 & 0.208 \\
 & 25\% & 0.170 & 0.196 & 0.150 & 0.200 & 0.166 & 0.192 & 0.201 & 0.201 \\
 & 10\% & 0.170 & 0.198 & 0.138 & 0.190 & 0.169 & 0.198 & 0.205 & 0.205 \\
 & 5\% & 0.176 & 0.202 & 0.139 & 0.190 & 0.177 & 0.199 & 0.215 & 0.224 \\ \hline
\multirow{4}{*}{Anecdotes} & 50\% & 0.271 & 0.293 & 0.227 & 0.295 & 0.285 & 0.316 & 0.301 & 0.268 \\
 & 25\% & 0.272 & 0.294 & 0.208 & 0.281 & 0.285 & 0.317 & 0.322 & 0.283 \\
 & 10\% & 0.271 & 0.294 & 0.212 & 0.287 & 0.303 & 0.322 & 0.300 & 0.271 \\
 & 5\% & 0.287 & 0.306 & 0.246 & 0.310 & 0.311 & 0.330 & 0.303 & 0.276 \\ \hline
\multirow{4}{*}{DynaSent} & 50\% & 0.264 & 0.305 & 0.211 & 0.296 & 0.299 & 0.349 & 0.279 & 0.264 \\
 & 25\% & 0.281 & 0.314 & 0.270 & 0.330 & 0.285 & 0.341 & 0.287 & 0.265 \\
 & 10\% & 0.263 & 0.305 & 0.207 & 0.294 & 0.285 & 0.341 & 0.296 & 0.275 \\
 & 5\% & 0.264 & 0.304 & 0.198 & 0.286 & 0.281 & 0.336 & 0.312 & 0.288 \\ \hline
\end{tabular}
} % end resizebox
\caption{Ablation on percentage of data used to train NAPHA. Results shown for base model SLP-SE.}
\label{tab:ablation_trainingData}
\end{table*}

\begin{table*}[]
\centering
\resizebox{\linewidth}{!}{%
\begin{tabular}{lc|cc|ccll|cc}
\cline{1-10}
                             & Alignment Model    & \multicolumn{2}{c|}{Linear Transformation} & \multicolumn{2}{c|}{Temperature Scaling} & \multicolumn{2}{l|}{Param. Temp. Scaling} & \multicolumn{2}{c}{Dirichlet} \\
                             & Alignment         & \multicolumn{2}{c|}{NAPHA}                 & \multicolumn{2}{c|}{NAPHA}                & \multicolumn{2}{c|}{NAPHA}                            & \multicolumn{2}{c}{NAPHA}     \\
                             & Entropy Labels      & predicted             & oracle             & predicted  & \multicolumn{1}{c|}{oracle} & predicted                   & oracle                  & predicted       & oracle      \\ \hline
\multirow{2}{*}{Anecdotes}   & DistCE $\downarrow$ & 0.285                 & 0.191              & 0.365      & \multicolumn{1}{c|}{0.265}  & 0.310                       & 0.320                   & 0.267           & 0.176       \\
                             & JSD $\downarrow$    & 0.302                 & 0.238              & 0.366      & \multicolumn{1}{c|}{0.300}  & 0.347                       & 0.336                   & 0.291           & 0.228       \\ \hline
\multirow{2}{*}{ChaosNLI}    & DistCE $\downarrow$ & 0.192                 & 0.182              & 0.232      & \multicolumn{1}{c|}{0.199}  & 0.215                       & 0.216                   & 0.171           & 0.149       \\
                             & JSD $\downarrow$    & 0.217                 & 0.210              & 0.244      & \multicolumn{1}{c|}{0.226}  & 0.231                       & 0.231                   & 0.197           & 0.184       \\ \hline
\multirow{2}{*}{DynaSent}    & DistCE $\downarrow$ & 0.281                 & 0.216              & 0.341      & \multicolumn{1}{c|}{0.225}  & 0.277                       & 0.286                   & 0.265           & 0.168       \\
                             & JSD $\downarrow$    & 0.317                 & 0.279              & 0.353      & \multicolumn{1}{c|}{0.275}  & 0.323                       & 0.330                   & 0.305           & 0.244       \\ \hline
\multirow{2}{*}{SummEval}    & DistCE $\downarrow$ & 0.319                 & 0.274              & 0.386      & \multicolumn{1}{c|}{0.411}  & 0.386                       & 0.375                   & 0.306           & 0.257       \\
                             & JSD $\downarrow$    & 0.368                 & 0.339              & 0.434      & \multicolumn{1}{c|}{0.444}  & 0.448                       & 0.442                   & 0.360           & 0.333       \\ \hline
\multirow{2}{*}{TopicalChat} & DistCE $\downarrow$ & 0.357                 & 0.322              & 0.357      & \multicolumn{1}{c|}{0.322}  & 0.319                       & 0.322                   & 0.357           & 0.322       \\
                             & JSD $\downarrow$    & 0.383                 & 0.366              & 0.383      & \multicolumn{1}{c|}{0.366}  & 0.366                       & 0.380                   & 0.383           & 0.366       \\ \hline
\end{tabular}
} % end resizebox
\caption{Ablation study on using different alignment model architectures. Results shown for base model SLP-SE.}
\label{tab:ablation_otherCalibrationModels}
\end{table*}

\section{Leave-One-Out Bootstrapping}\label{appdx:leaveOneOut}
To estimate the human performance, we modify the bootstrapping approach from \citet{bavaresco-etal-2025-llms} to a leave-one-out bootstrapping.
Given the small number (3) of annotators for these two datasets, we need to measure with the leave-one-out mechanism. Otherwise, the estimation of human performance would be too optimistic, since $\frac{1}{3}$ of the value that a single annotator is being compared to, is his own judgment, introducing a statistical bias.
We average the results of 1000 iterations to estimate the human performance.
In each iteration, one annotator is randomly picked. This single annotator judgment is then compared to the aggregated judgment from the other two annotators. In our case, we calculate the Kendall's $\tau$ correlation.
This process is similar to how the alt-test \cite{calderon-etal-2025-alternative} estimates the human performance.

\section{Results with other Backbone LLMs} \label{sec:appdx_otherModels}
While our results mainly focus on using Claude-4-Sonnet as a backbone LLM, we also report results with GPT-OSS-120B~\cite{agarwal2025gpt} and Qwen3-32B \cite{yang2025qwen3}.

For the \emph{hard-label} predictions, the results are shown in Tab. \ref{tab:f1_scores_otherBackbones}.
The soft-label predictions are shown in Tables \ref{tab:results_otherBackbone_soft_anecdotes} to \ref{tab:results_otherBackbones_soft_topical}.

\begin{table}[tbh]
\centering
\resizebox{\linewidth}{!}{%% \usepackage{multirow}
\begin{tabular}{llccc}
\hline
Base Model & Dataset / Entropy class & Low $H$ & Medium $H$ & High $H$ \\ \hline
\multirow{3}{*}{GPT-OSS-120B} & ChaosNLI & 0.93 & 0.77 & 0.56 \\
 & Anecdotes & 0.35 & 0.38 & 0.28 \\
 & DynaSent & 0.96 & 0.81 & 0.39 \\ \hline
\multirow{3}{*}{Qwen3-32B} & ChaosNLI & 0.92 & 0.74 & 0.59 \\
 & Anecdotes & 0.66 & 0.41 & 0.22 \\
 & DynaSent & 0.96 & 0.80 & 0.40 \\ \hline
\end{tabular}
} % end resizebox
\caption{Results from \emph{hard-label} predictions from GPT-OSS-120B and Qwen3-32B, stratified by entropy class. Reporting macro-Average F1 Scores.}
\label{tab:f1_scores_otherBackbones}
\end{table}

\begin{table*}[tbh]
\centering
\resizebox{\linewidth}{!}{%
\begin{tabular}{llcccc|cccccc}
\hline
 &  &  &  & \multicolumn{2}{c|}{Average} & \multicolumn{2}{c}{Low $H$} & \multicolumn{2}{c}{Medium $H$} & \multicolumn{2}{c}{High $H$} \\
Backbone & Base model & Alignment & Entropy Labels & DistCE $\downarrow$ & JSD $\downarrow$ & DistCE $\downarrow$ & JSD $\downarrow$ & DistCE $\downarrow$ & JSD $\downarrow$ & DistCE $\downarrow$ & JSD $\downarrow$ \\ \hline
\multirow{6}{*}{GPT-OSS-120B} & \multirow{3}{*}{SLP-SE} & none & none & 0.365 & 0.376 & 0.289 & 0.363 & 0.376 & 0.392 & 0.431 & 0.372 \\
 &  & NAPHA & predicted & $0.321$ \small{(0.001)} & $0.327$ \small{(0.001)} & $0.331$ \small{(0.004)} & $0.367$ \small{(0.002)} & $0.325$ \small{(0.002)} & $0.333$ \small{(0.001)} & $0.305$ \small{(0.003)} & $0.273$ \small{(0.002)} \\
 &  & NAPHA & oracle & $0.222$ \small{(0.002)} & $0.265$ \small{(0.001)} & $0.140$ \small{(0.005)} & $0.247$ \small{(0.002)} & $0.314$ \small{(0.002)} & $0.325$ \small{(0.001)} & $0.210$ \small{(0.000)} & $0.210$ \small{(0.000)} \\ \cline{2-12} 
 & \multirow{3}{*}{SimAnn} & none & none & 0.374 & 0.420 & 0.150 & 0.281 & 0.410 & 0.437 & 0.564 & 0.508 \\
 &  & NAPHA & predicted & $0.315$ \small{(0.001)} & $0.325$ \small{(0.001)} & $0.322$ \small{(0.003)} & $0.361$ \small{(0.002)} & $0.302$ \small{(0.001)} & $0.325$ \small{(0.001)} & $0.321$ \small{(0.003)} & $0.285$ \small{(0.002)} \\
 &  & NAPHA & oracle & $0.205$ \small{(0.001)} & $0.254$ \small{(0.001)} & $0.109$ \small{(0.003)} & $0.222$ \small{(0.002)} & $0.296$ \small{(0.002)} & $0.317$ \small{(0.002)} & $0.210$ \small{(0.001)} & $0.210$ \small{(0.001)} \\ \hline
\multirow{6}{*}{Qwen3-32B} & \multirow{3}{*}{SLP-SE} & none & none & 0.3910 & 0.396  & 0.397  & 0.424 & 0.402 & 0.408 & 0.373  & 0.351  \\
 &  & NAPHA & predicted & $0.333$ {(0.001)} & $0.336$ {(0.001)} & $0.364$ {(0.003)} & $0.387$ {(0.002)} & $0.330$ {(0.001)} & $0.339$ {(0.001)} & $0.304$ {(0.002)} & $0.272$ {(0.002)} \\
 &  & NAPHA & oracle& $0.237$ {(0.002)} & $0.278$ {(0.001)} & $0.164$ {(0.005)} & $0.271$ {(0.002)} & $0.333$ {(0.001)} & $0.336$ {(0.001)} & $0.213$ {(0.000)} & $0.214$ {(0.000)} \\ \cline{2-12} 
 & \multirow{3}{*}{SimAnn} & none & none & 0.533 & 0.515 & 0.566 & 0.566 & 0.534 & 0.508 & 0.499 & 0.467 \\
 &  & NAPHA & predicted & $0.359$ {(0.001)} & $0.369$ {(0.001)} & $0.351$ {(0.005)} & $0.386$ {(0.003)} & $0.369$ {(0.002)} & $0.374$ {(0.002)} & $0.359$ {(0.004)} & $0.346$ {(0.003)} \\
 &  & NAPHA & oracle & $0.351$ {(0.001)} & $0.347$ {(0.001)} & $0.402$ {(0.004)} & $0.409$ {(0.002)} & $0.337$ {(0.001)} & $0.342$ {(0.001)} & $0.315$ {(0.004)} & $0.280$ {(0.003)} \\ \hline
\end{tabular}
} % end resizebox
\caption{Results Table Anecdotes soft-labels with GPT-OSS-120B and Qwen3-32B backbones.} 
\label{tab:results_otherBackbone_soft_anecdotes}
\end{table*}

\begin{table*}[tbh]
\centering
\resizebox{\linewidth}{!}{%
\begin{tabular}{llcccc|cccccc}
\hline
 &  &  &  & \multicolumn{2}{c|}{Average} & \multicolumn{2}{c}{Low $H$} & \multicolumn{2}{c}{Medium $H$} & \multicolumn{2}{c}{High $H$} \\
Backbone & Base model & Alignment & Entropy Labels & DistCE $\downarrow$ & JSD $\downarrow$ & DistCE $\downarrow$ & JSD $\downarrow$ & DistCE $\downarrow$ & JSD $\downarrow$ & DistCE $\downarrow$ & JSD $\downarrow$ \\ \hline
\multirow{6}{*}{GPT-OSS-120B} & \multirow{3}{*}{SLP-SE} & none & none & 0.222 & 0.256 & 0.094 & 0.188 & 0.230 & 0.261 & 0.338 & 0.301 \\
 &  & NAPHA & predicted & $0.201$ \small{(0.001)} & $0.220$ \small{(0.001)} & $0.201$ \small{(0.003)} & $0.236$ \small{(0.002)} & $0.191$ \small{(0.001)} & $0.213$ \small{(0.001)} & $0.229$ \small{(0.002)} & $0.221$ \small{(0.001)} \\
 &  & NAPHA & oracle & $0.171$ \small{(0.001)} & $0.202$ \small{(0.001)} & $0.103$ \small{(0.003)} & $0.182$ \small{(0.003)} & $0.189$ \small{(0.002)} & $0.211$ \small{(0.002)} & $0.191$ \small{(0.001)} & $0.196$ \small{(0.001)} \\ \cline{2-12} 
 & \multirow{3}{*}{SimAnn} & none & none & 0.260 & 0.313 & 0.092 & 0.187 & 0.275 & 0.320 & 0.397 & 0.390 \\
 &  & NAPHA & predicted & $0.205$ \small{(0.001)} & $0.230$ \small{(0.001)} & $0.197$ \small{(0.003)} & $0.234$ \small{(0.002)} & $0.200$ \small{(0.001)} & $0.230$ \small{(0.001)} & $0.230$ \small{(0.002)} & $0.226$ \small{(0.001)} \\
 &  & NAPHA & oracle & $0.168$ \small{(0.001)} & $0.205$ \small{(0.001)} & $0.067$ \small{(0.004)} & $0.147$ \small{(0.004)} & $0.198$ \small{(0.001)} & $0.226$ \small{(0.001)} & $0.187$ \small{(0.001)} & $0.195$ \small{(0.001)} \\ \hline
 
\multirow{6}{*}{Qwen3-32B} & \multirow{3}{*}{SLP-SE} & none & none & 0.319 & 0.323 & 0.207 & 0.279 & 0.342 & 0.334 & 0.374 & 0.333  \\
 &  & NAPHA & predicted & $0.216$ {(0.001)} & $0.227$ {(0.001)} & $0.240$ {(0.004)} & $0.263$ {(0.003)} & $0.208$ {(0.001)} & $0.215$ {(0.001)} & $0.212$ {(0.002)} & $0.216$ {(0.002)} \\
 &  & NAPHA & oracle & $0.195$ {(0.001)} & $0.209$ {(0.001)} & $0.144$ {(0.004)} & $0.215$ {(0.003)} & $0.215$ {(0.002)} & $0.211$ {(0.002)} & $0.193$ {(0.002)} & $0.196$ {(0.001)} \\ \cline{2-12} 
 & \multirow{3}{*}{SimAnn} & none & none & 0.224 & 0.271 & 0.102 & 0.192 & 0.236 & 0.276 & 0.319 & 0.324 \\
 &  & NAPHA & predicted & $0.206$ {(0.001)} & $0.224$ {(0.001)} & $0.226$ {(0.004)} & $0.256$ {(0.002)} & $0.195$ {(0.002)} & $0.216$ {(0.001)} & $0.216$ {(0.002)} & $0.211$ {(0.002)} \\
 &  & NAPHA & oracle & $0.173$ {(0.002)} & $0.201$ {(0.001)} & $0.106$ {(0.006)} & $0.185$ {(0.004)} & $0.194$ {(0.002)} & $0.210$ {(0.002)} & $0.184$ {(0.001)} & $0.188$ {(0.001)} \\ \hline
\end{tabular}
} % end resizebox
\caption{Results Table ChaosNLI soft-labels with GPT-OSS-120B and Qwen3-32B backbones.}
\label{tab:results_otherBackbone_soft_nli}
\end{table*}

\begin{table*}[tbh]
\centering
\resizebox{\linewidth}{!}{%
\begin{tabular}{llcccc|cccccc}
\hline
 &  &  &  & \multicolumn{2}{c|}{Average} & \multicolumn{2}{c}{Low $H$} & \multicolumn{2}{c}{Medium $H$} & \multicolumn{2}{c}{High $H$} \\ \hline
Backbone & Base model & Alignment & Entropy Labels & DistCE $\downarrow$ & JSD $\downarrow$ & DistCE $\downarrow$ & JSD $\downarrow$ & DistCE $\downarrow$ & JSD $\downarrow$ & DistCE $\downarrow$ & JSD $\downarrow$ \\ \hline
\multirow{6}{*}{GPT-OSS-120B} & \multirow{3}{*}{SLP-SE} & none & none & 0.277 & 0.317 & 0.133 & 0.234 & 0.256 & 0.321 & 0.443 & 0.378 \\
 &  & NAPHA & predicted & $0.288$ \small{(0.001)} & $0.322$ \small{(0.001)} & $0.268$ \small{(0.006)} & $0.333$ \small{(0.004)} & $0.307$ \small{(0.002)} & $0.355$ \small{(0.001)} & $0.288$ \small{(0.005)} & $0.273$ \small{(0.003)} \\
 &  & NAPHA & oracle & $0.175$ \small{(0.003)} & $0.250$ \small{(0.002)} & $0.067$ \small{(0.008)} & $0.175$ \small{(0.007)} & $0.277$ \small{(0.006)} & $0.338$ \small{(0.003)} & $0.179$ \small{(0.001)} & $0.204$ \small{(0.000)} \\ \cline{2-12} 
 & \multirow{3}{*}{SimAnn} & none & none & 0.304 & 0.375 & 0.035 & 0.142 & 0.277 & 0.346 & 0.600 & 0.532 \\
 &  & NAPHA & predicted & $0.305$ \small{(0.002)} & $0.330$ \small{(0.001)} & $0.293$ \small{(0.007)} & $0.345$ \small{(0.004)} & $0.280$ \small{(0.003)} & $0.341$ \small{(0.002)} & $0.341$ \small{(0.004)} & $0.301$ \small{(0.003)} \\
 &  & NAPHA & oracle & $0.165$ \small{(0.004)} & $0.242$ \small{(0.003)} & $0.063$ \small{(0.010)} & $0.173$ \small{(0.008)} & $0.250$ \small{(0.005)} & $0.322$ \small{(0.003)} & $0.181$ \small{(0.001)} & $0.204$ \small{(0.001)} \\ \hline
\multirow{6}{*}{Qwen3-32B} & \multirow{3}{*}{SLP-SE} & none & none & 0.277 & 0.317 & 0.133 & 0.234 & 0.256 & 0.321 & 0.443 & 0.378 \\
 &  & NAPHA & predicted & $0.288$ {(0.001)} & $0.322$ {(0.001)} & $0.268$ {(0.006)} & $0.333$ {(0.004)} & $0.307$ {(0.002)} & $0.355$ {(0.001)} & $0.288$ {(0.005)} & $0.273$ {(0.003)} \\
 &  & NAPHA & oracle & $0.175$ {(0.003)} & $0.250$ {(0.002)} & $0.067$ {(0.008)} & $0.175$ {(0.007)} & $0.277$ {(0.006)} & $0.338$ {(0.003)} & $0.179$ {(0.001)} & $0.204$ {(0.000)} \\ \cline{2-12} 
 & \multirow{3}{*}{SimAnn} & none & none & 0.291 & 0.359 & 0.037 & 0.134 & 0.281 & 0.346 & 0.553 & 0.499 \\
 &  & NAPHA & predicted & $0.293$ {(0.001)} & $0.321$ {(0.001)} & $0.269$ {(0.006)} & $0.330$ {(0.004)} & $0.281$ {(0.002)} & $0.340$ {(0.001)} & $0.329$ {(0.004)} & $0.293$ {(0.003)} \\
 &  & NAPHA & oracle & $0.167$ {(0.003)} & $0.245$ {(0.002)} & $0.058$ {(0.008)} & $0.166$ {(0.008)} & $0.267$ {(0.005)} & $0.332$ {(0.003)} & $0.176$ {(0.002)} & $0.203$ {(0.001)} \\\hline
\end{tabular}
} % end resizebox
\caption{Results Table DynaSent soft-labels with GPT-OSS-120B and Qwen3-32B backbones.}
\label{tab:results_otherBackbones_soft_sentiment}
\end{table*}

\begin{table*}[tbh]
\centering
\resizebox{0.8\linewidth}{!}{%
\begin{tabular}{llcccc}
\hline
 &  &  &  & \multicolumn{2}{c}{Average} \\
Backbone & Base model & Alignment & Entropy Labels & DistCE $\downarrow$ & JSD $\downarrow$ \\ \hline
\multirow{6}{*}{GPT-OSS-120B} & \multirow{3}{*}{SLP-SE} & none & none & 0.482 & 0.506 \\
 &  & NAPHA & predicted & 0.330 & 0.379 \\
 &  & NAPHA & oracle & 0.282 & 0.349 \\ \cline{2-6} 
 & \multirow{3}{*}{SimAnn} & none & none & 0.524 & 0.530 \\
 &  & NAPHA & predicted & 0.309 & 0.362 \\
 &  & NAPHA & oracle & 0.264 & 0.334 \\ \hline
\multirow{6}{*}{Qwen3-32B} & \multirow{3}{*}{SLP-SE} & none & none & 0.338 & 0.424 \\
 &  & NAPHA & predicted & none$^\ast$ & none$^\ast$ \\
 &  & NAPHA & oracle & 0.297 & 0.359 \\ \cline{2-6} 
 & \multirow{3}{*}{SimAnn} & none & none & 0.469 & 0.501 \\
 &  & NAPHA & predicted & 0.307 & 0.361 \\
 &  & NAPHA & oracle & 0.260 & 0.332 \\ \hline
\end{tabular}
} % end resizebox
\caption{Results Table SummEval soft-labels with GPT-OSS-120B and Qwen3-32B backbones.$^\ast$: Prediction did not allow for train-test split stratified by entropy-level. Displaying the average over the evaluation dimensions}
\label{tab:results_otherBackbones_soft_sumeval}
\end{table*}

\begin{table*}[tbh]
\centering
\resizebox{0.8\linewidth}{!}{%
\begin{tabular}{llcccc}
\hline
 &  &  &  & \multicolumn{2}{c}{Average} \\
Backbone & Base model & Alignment & Entropy Labels & DistCE $\downarrow$ & JSD $\downarrow$ \\ \hline
\multirow{6}{*}{GPT-OSS-120B} & \multirow{3}{*}{SLP-SE} & none & none & 0.355 & 0.379 \\
 &  & NAPHA & predicted & none$^\ast$ & none$^\ast$ \\
 &  & NAPHA & oracle & 0.244 & 0.309 \\ \cline{2-6} 
 & \multirow{3}{*}{SimAnn} & none & none & 0.365 & 0.409 \\
 &  & NAPHA & predicted & 0.301 & 0.338 \\
 &  & NAPHA & oracle & 0.225 & 0.295 \\ \hline
\multirow{6}{*}{Qwen3-32B} & \multirow{3}{*}{SLP-SE} & none & none & 0.405 & 0.472 \\
 &  & NAPHA & predicted &  none$^\ast$ & none$^\ast$ \\
 &  & NAPHA & oracle & 0.369 & 0.400 \\ \cline{2-6} 
 & \multirow{3}{*}{SimAnn} & none & none & 0.304 & 0.358 \\
 &  & NAPHA & predicted & 0.312 & 0.347 \\
 &  & NAPHA & oracle & 0.236 & 0.306 \\ \hline
\end{tabular}
} % end resizebox
\caption{Results Table TopicalChat soft-labels with GPT-OSS-120B and Qwen3-32B backbones.$^\ast$: Prediction did not allow for train-test split stratified by entropy-level. Displaying the average over the evaluation dimensions}
\label{tab:results_otherBackbones_soft_topical}
\end{table*}

\section{Case Studies} \label{sec:appdx_caseStudy}
We present two case studies and their discussion in Figures \ref{fig:case-study-1} and \ref{fig:case-study-2}.

\casestudy
    {Case study 1: Example from the Anecdotes dataset, discussing pricing for commissioned work to family.}
    {\textbf{Title:} AITA for expecting one of my family members to pay for my painting? \textbf{Situation:} My uncle
  wants a portrait done of his sons. Painting is hard and very time consuming when you take it seriously. As much as I am
  passionate about it, when I paint something I wasn't motivate to do in the first place, I need a motivation and that is
  commission! So I tell my uncle I will charge him the rate I charge my other clients. He gets upset and insists I do it
  for free because it's a family thing. I'm a college student. Painting supplies are expensive, not to mention tuition.
  And most importantly, my time is valuable during this phase of my life.}
    {The author is a college student who paints and charges clients for commissioned work. Their uncle wants a portrait of
  his sons and expects it to be done for free because they are family. The author wants to charge their normal rate,
  citing that painting supplies are expensive, they need money for tuition, and their time is valuable.
  
    Key considerations:
    \begin{itemize}
      \item The author has established rates for their painting services
      \item They are a college student with financial needs (tuition, supplies)
      \item The uncle is requesting a service that normally costs money
      \item The uncle expects free work solely based on family relationship
      \item The author treats this as their business/source of income
    \end{itemize}
  
    The author is not wrong for:
    \begin{itemize}
      \item Wanting to be paid for their skilled work and time
      \item Treating family clients the same as other clients professionally
      \item Needing income as a college student
      \item Valuing their artistic services appropriately
    \end{itemize}
  
    The uncle is wrong for:
    \begin{itemize}
      \item Expecting free professional services just because of family relationship
      \item Getting upset when asked to pay normal rates
      \item Not respecting the author's time, skill, and financial needs
      \item Pressuring family to work for free
    \end{itemize}
  
    This is a clear case where the uncle is trying to exploit the family relationship to get free professional services,
  while the author is reasonably asking to be compensated for their work.
  
    \texttt{[AUTHOR]: 0.05 \quad [OTHER]: 0.90 \quad [EVERYBODY]: 0.02 \quad [NOBODY]: 0.02 \quad [INFO]: 0.01}}
    {The LLM makes a clear decision: the uncle ([OTHER]), not the student ([AUTHOR]). It takes the stance that ``The author
  is not wrong for: Treating family clients the same as other clients professionally.'' This aligns with a more
  individualist cultural perspective. Importantly, the reasoning trace completely lacks consideration of other
  perspectives and cultural norms. E.g., in more collectivist cultures, providing services to family free of charge is
  more expected and highly valued. In this specific example, the LLM prediction actually aligns with the human ground-truth. The
  inherent model bias happens to match the human majority perspective. Nevertheless, the reasoning trace shows the model
  is not considering other perspectives.}
  {fig:case-study-1}

\casestudy
    {Case study: differing communication expectations about sharing news (AITA).}
    {\textbf{Title:} AITA for being upset my girlfriend chooses to share things with others before me? \textbf{Situation:}
  I (24M) have a girlfriend (21F). She's in college studying graphic design, and she recently got a job at the school's
  newspaper. She designed the front page layout and created illustrations. This was a big deal, (at least in my eyes). She
  wants to be a graphic designer, and this is directly related to that. I am proud of her for doing it, but I only found
  out about it through social media. This is not the first time things like this have happened. I think it boils down to
  of it were me, I would be so excited to share this news with my significant other. My first action would not be to post
  it on social media to let everyone else know first. So, AITA for wanting to be the first person my SO tells news to? Or
  is it her news and she can tell who she chooses?}
    {Let me carefully analyze this situation to determine who is in the wrong. The author is upset that his girlfriend
  shares her achievements and news on social media before telling him directly. He feels that as her significant other, he
  should be the first person she tells important news to, rather than finding out through social media posts. Let me
  consider each party's position:
  
    The girlfriend's perspective:
    \begin{itemize}
      \item She achieved something she's proud of (getting a job at the school newspaper and designing the front page)
      \item She chose to share this news on social media first
      \item She has the right to decide how and when to share her own news
      \item There's no indication she was intentionally trying to hurt or exclude her boyfriend
    \end{itemize}
  
    The author's perspective:
    \begin{itemize}
      \item He feels hurt that he wasn't told first about his girlfriend's achievement
      \item He believes significant others should share important news with each other before posting on social media
      \item He's expressed this is a pattern of behavior that bothers him
    \end{itemize}
  
    The key question is: Is either party doing something wrong? The girlfriend is not obligated to follow a specific order
  when sharing her own news. While it might be considerate to tell a significant other first, it's not a requirement, and
  she's not doing anything inherently wrong by posting on social media first.
  
    The author, while his feelings are understandable, is trying to impose expectations on how his girlfriend should share
  her own achievements. His desire to be told first, while natural, doesn't make his girlfriend wrong for not doing so.
  
    This appears to be a case where both parties have valid perspectives, but neither is actually doing something wrong.
  It's more about different communication styles and expectations in the relationship.
  
    \texttt{[AUTHOR]: 0.15 \quad [OTHER]: 0.05 \quad [EVERYBODY]: 0.05 \quad [NOBODY]: 0.70 \quad [INFO]: 0.05}}
    {In the second case study, the LLM argues that nobody is wrong, since the situation reflects different communication
  styles. However, the ground-truth is very different: roughly 2/3 see the author in the wrong, and 1/3 the other party.
  The LLM gravitates toward avoiding to assign blame to any of the two parties, providing a conflict-averse response. The
  human opinions are much more polarized.}
  {fig:case-study-2}

\section{Prompts} \label{sec:appdx_prompts}
\begin{figure}
    \centering
    \includegraphics[width=1\linewidth]{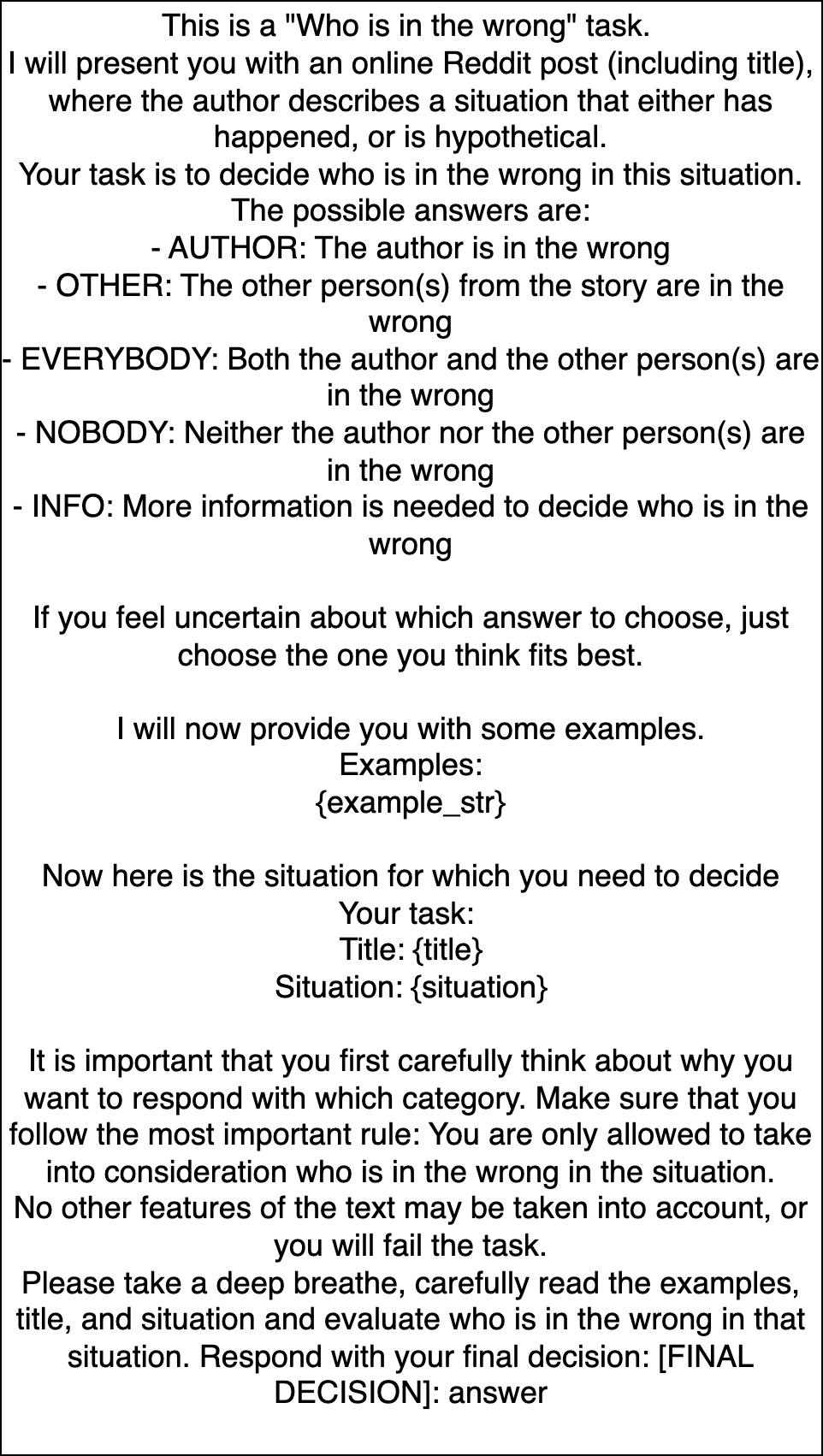}
    \caption{Example prompt for the \emph{hard-label} base model prediction. Shown for the Anecdotes dataset}
    \label{fig:prompt_hardLabel}
\end{figure}

\begin{figure}
    \centering
    \includegraphics[width=1\linewidth]{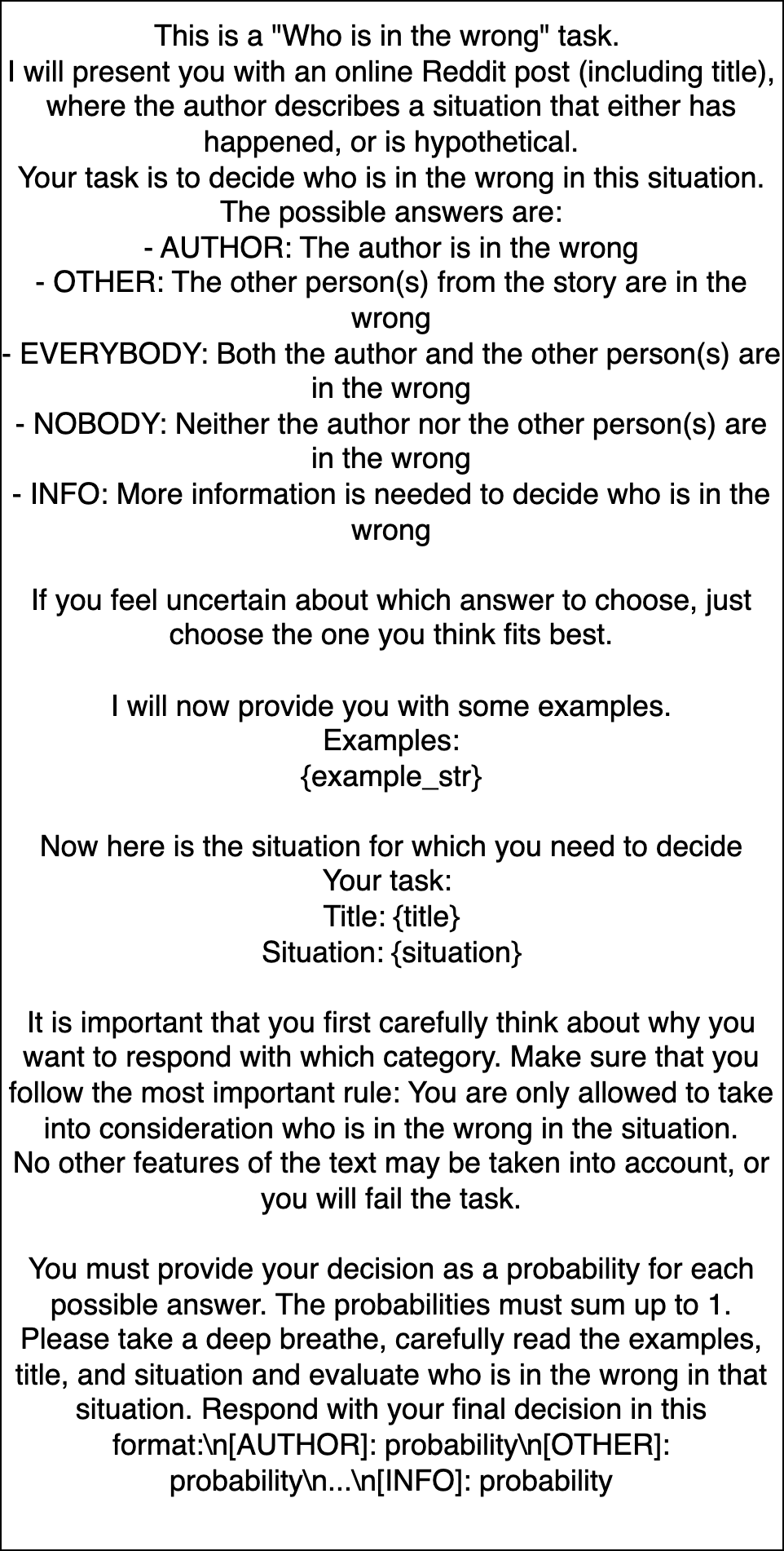}
    \caption{Example prompt for the soft-label base model prediction. Shown for the Anecdotes dataset}
    \label{fig:prompt_softLabel}
\end{figure}

Figures \ref{fig:prompt_hardLabel} and \ref{fig:prompt_softLabel} showcase example prompts for hard- and soft-label prediction, respectively. These were used for the Anecdotes dataset. The prompts for the other datasets follow the same structure and only minimal changes were made to fit the specific task.

\section{AI Assistant Usage}
We used AI assistants as coding assistance and for basic proof-reading of our manuscript. We are solely responsible for all content.  

\end{document}